\documentclass{article} 
\usepackage[final]{colm2026_conference}

\usepackage{microtype}
\usepackage{hyperref}
\usepackage{url}

\usepackage{lineno}

\definecolor{darkblue}{rgb}{0, 0, 0.5}
\hypersetup{colorlinks=true, citecolor=darkblue, linkcolor=darkblue, urlcolor=darkblue}

\usepackage{graphicx}
\usepackage{cuted}
\usepackage{capt-of}
\usepackage{amsmath}
\usepackage{caption}
\usepackage{subcaption}
\usepackage{amssymb}
\usepackage{multibib}
\usepackage{wrapfig}
\usepackage{multirow}
\usepackage{multicol}
\usepackage{booktabs}
\usepackage{enumitem}
\usepackage{pifont}
\usepackage{lipsum}
\usepackage{float}
\usepackage[table,dvipsnames]{xcolor}
\usepackage{colortbl}
\usepackage{tikz}
\usepackage{xurl}
\usepackage[most]{tcolorbox}
\usepackage{listings}

\usepackage{tabularx}
\usetikzlibrary{shapes.geometric, arrows.meta, positioning, calc, fit, backgrounds, patterns, matrix, shadows}

\newcommand{\cmark}{\textcolor{Green}{\ding{51}}}
\newcommand{\xmark}{\textcolor{Red}{\ding{55}}}

\DeclareRobustCommand{\ours}{\texorpdfstring{\textsc{HyperThink}}{HyperThink}}

\lstdefinestyle{llmblock}{
  basicstyle=\ttfamily\fontsize{8.6}{4}\selectfont,
  columns=fullflexible,
  breaklines=true,
  breakatwhitespace=true,
  keepspaces=true,
  showstringspaces=false,
  upquote=true,
  tabsize=2,
  resetmargins=true,
  xleftmargin=0pt,
  xrightmargin=0pt,
  breakindent=0pt,
  breakautoindent=false,
}

\newtcolorbox{InteractionBox}[1]{
  enhanced,
  breakable,
  colback=blue!2,
  colframe=blue!55!black,
  colbacktitle=blue!10,
  coltitle=blue!60!black,
  boxrule=0.6pt,
  arc=2pt,
  left=6pt,right=6pt,top=6pt,bottom=6pt,
  title={#1},
  fonttitle=\bfseries,
}

\newcommand{\FieldHeader}[1]{\par\smallskip\noindent\textbf{#1}\par\smallskip}

\lstnewenvironment{LLMBlock}[1][]%
{\vspace{-2pt}\lstset{style=llmblock,#1}}%
{\vspace{-2pt}}

\usepackage{titlesec}
\titlespacing{\section}{0pt}{-1.5pt}{-2pt}
\titlespacing{\subsection}{0pt}{-2pt}{-2pt}
\titlespacing{\paragraph}{0pt}{0pt}{0.3\baselineskip}
\setlist{nosep, topsep=2pt, partopsep=2pt, itemsep=2pt, parsep=2pt}
\setlist[itemize,enumerate]{leftmargin=*, labelsep=0.4em}

\title{\ours{}: Text-to-Parameter Hypernetworks for Efficient Reasoning}

\author{
\textbf{Donggyun Kim}$^1$\thanks{equal contribution}\quad
\textbf{Jack Lu}$^2$\footnotemark[1]\quad
\textbf{Chanwoo Kim}$^1$\quad
\textbf{Mengye Ren}$^2$\quad
\textbf{Seunghoon Hong}$^1$\\[0.15cm]
$^1$ KAIST School of Computing\\
$^2$ NYU Courant Institute School of Mathematics, Computing, and Data Science\\[0.15cm]
\hspace{0.1cm}{\small\ttfamily
\begin{tabular}{@{}l@{}}
donggyun.kim@kaist.ac.kr, yl11330@nyu.edu, kcwkcw0802@kaist.ac.kr\\
mengye@nyu.edu, seunghoon.hong@kaist.ac.kr
\end{tabular}
}
}

\begin{document}

\ifcolmsubmission
\linenumbers
\fi

\maketitle

\begin{abstract}
Long-form thinking traces can substantially improve the multi-step reasoning performance of large language models (LLMs), but they introduce high inference-time overhead, with latency dominated by sequential decoding.
We propose \ours{}, a text-to-parameter approach that amortizes this reasoning computation into a single query-conditioned parameter update: a lightweight hypernetwork reads the question and predicts updates to a small subset of the base LLM’s parameters, while a vector-quantized decoder constrains them to a finite set of reusable patterns to improve robustness and transfer.
Trained end-to-end on outputs from the base model itself, \ours{} eliminates long thinking traces at test time: after one hypernetwork forward pass, the adapted model generates a concise step-by-step solution and final answer without an intermediate trace, using far fewer tokens while retaining strong reasoning performance.
Empirically, \ours{} improves the low-latency region of the accuracy--latency trade-off on mathematical and general reasoning tasks, with its strongest gains in the near-non-thinking regime.
\end{abstract}

\vspace{0.3cm}
\section{Introduction}
\label{sec:introduction}

Large language models (LLMs) have rapidly advanced on tasks requiring multi-step inference, including mathematical and symbolic reasoning~\citep{openai2024o1, llama3, qwen3}.
A key driver of this progress is chain-of-thought (CoT) reasoning, where models generate \emph{step-by-step responses} before giving a final answer~\citep{wei2022chain, kojima2022large}.
Beyond CoT prompting, recent reasoning models are also post-trained to produce long \emph{thinking traces} before these step-by-step responses, typically enclosed by special tokens such as {\footnotesize \texttt{<think>} ... \texttt{</think>}}~\citep{qwen3, deepseekr1, olmo3}.
Although effective, such traces are computationally expensive at inference time because they often dominate the token budget, making them impractical in latency-sensitive or cost-constrained settings.

To address this cost, recent work has begun exploring \emph{non-thinking} inference as an alternative, aiming to generate direct answers without intermediate reasoning.
Some methods incorporate \emph{non-thinking} behavior directly into post-trained reasoning models by providing explicit controls to disable thinking traces at test time~\citep{qwen3, fang2025llm}.
Others show that even existing reasoning models can often bypass explicit thinking prompts while retaining favorable accuracy--latency tradeoffs~\citep{ma2025reasoning}.
However, these approaches inherit a fundamental limitation of \emph{non-thinking}: on multi-step problems, models may produce plausible-looking answers through shortcuts rather than reliably executing the intermediate computation needed for correctness~\citep{wei2022chain}, leading to subpar performance.
This raises a central question: \emph{can we preserve much of the benefit of explicit thinking traces while operating at, or close to, the cost of decoding only short step-by-step responses?}

We approach this question by asking what makes thinking traces and CoT effective in the first place: they enable \emph{query-dependent computation}, allowing the model to adapt its internal processing to each problem instance.
If token-level thinking is reduced or removed, an alternative mechanism must still provide instance-specific adaptation, rather than relying on a single static set of weights for all queries.
At the same time, parameter-efficient adaptation methods suggest a promising route: small, structured weight modifications, such as bias tuning or low-rank updates, can substantially steer model behavior while changing only a tiny fraction of parameters~\citep{hu2022lora, zaken2022bitfit, beft}.
Together, these observations motivate a different perspective: instead of treating reasoning as a thinking trace to be generated, we treat it as an \emph{instance-conditioned perturbation of the model weights}, predicted from the query and applied before decoding only a short step-by-step response.

\begin{wrapfigure}[27]{r}{0.5\textwidth}
    \centering
    \includegraphics[trim={0cm 0cm 0cm 2.5cm},clip,width=\linewidth]{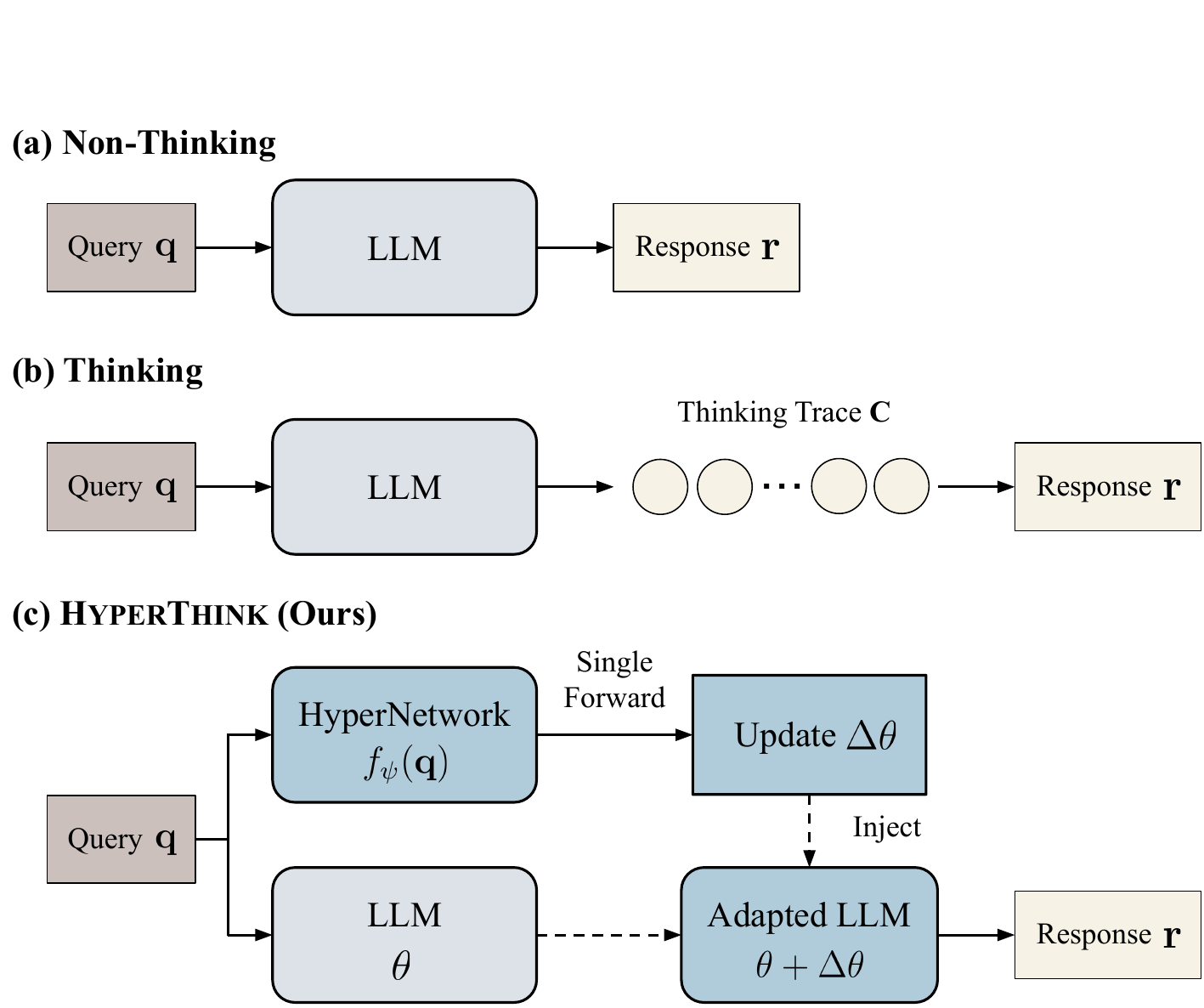}
    \vspace{0.05in}
    \caption{
        \looseness=-10000
        \textbf{Conceptual comparison.}
        (a) Non-Thinking mode produces the step-by-step response directly from the query.
        (b) Thinking mode produces a long thinking trace before producing the response, incurring high cost and latency.
        (c) \textbf{\ours{}} amortizes the reasoning computation into a query-conditioned parameter update $\Delta\theta$.
        The hypernetwork predicts this update in a single forward pass, enabling the adapted LLM to produce the response directly without a thinking trace.
    }
    \label{fig:concept_figure}
\end{wrapfigure}

Concretely, we approximate the downstream effect of thinking with a \emph{query-conditioned parameter update}.
Rather than generating a long trace at inference time, we learn a mapping from the input question to a lightweight parameter modulation that shifts a frozen base model into a configuration better suited to solving that question with few tokens.
We instantiate this idea in \ours{}, a text-to-parameter framework that amortizes instance-specific reasoning into query-conditioned parameter updates applied to the base model (Figure~\ref{fig:concept_figure}).
Given a query, a hypernetwork predicts a small set of parameter shifts, after which the adapted model produces a concise step-by-step solution and final answer without emitting a long thinking trace.

To improve robustness and reuse across instances, \ours{} introduces a vector-quantized (VQ) bottleneck that constrains updates to a finite set of discrete prototypical patterns~\citep{van2017neural, esser2021taming}.
The hypernetwork is trained end-to-end using supervision from the \emph{thinking mode} of the base LLM: during training, it learns to match the responses produced when conditioning on the long thinking trace in context.
As a result, inference requires only a single hypernetwork forward pass on the input problem, followed by standard decoding of a concise step-by-step response.
Because the hypernetwork is lightweight, \ours{} achieves computational efficiency close to \emph{non-thinking} while producing more reliable responses.
Moreover, its query-conditioned modulation keeps the base LLM frozen and introduces only a small number of extra parameters, which is beneficial for modular and scalable adaptation.

Across reasoning benchmarks spanning mathematics, code generation, and question answering, \ours{} shows its clearest gains in the low-latency regime, particularly on mathematical and question-answering tasks.
Unlike token-level methods for efficient reasoning, which still allocate additional computation through generated reasoning tokens or budget-controlled traces, \ours{} uses a single non-autoregressive, query-conditioned parameter update.
This makes \ours{} particularly useful in the near-non-thinking regime, where native non-thinking is fast but can sacrifice accuracy, whereas full thinking is too slow for strict latency budgets.
Taken together, our results suggest that amortized, query-conditioned parameter steering can recover useful instance-specific reasoning behavior without requiring long explicit thinking traces.

\section{Preliminaries}
\label{sec:preliminaries}

\vspace{0.1cm}
\subsection{Large Language Models for Problem Solving}

For problem solving, large language models (LLMs) can be viewed as conditional autoregressive sequence models over a vocabulary $\mathcal{V}$. Given an input query $\mathbf{q} = (q_1, \dots, q_M) \in \mathcal{V}^M$, the model $p_\theta$ autoregressively generates an output response $\mathbf{r} = (r_1, \cdots, r_T) \in \mathcal{V}^T$:
\begin{equation}
    p_\theta(\mathbf{r}\mid\mathbf{q}) = \prod_{t=1}^T p_\theta(r_t\mid \mathbf{r}_{<t}, \mathbf{q}) .
\end{equation}
The learning objective is to estimate parameters $\theta$ such that the model assigns high probability to correct responses on unseen queries.

In this work, we focus on reasoning-based problem solving, where the response $\mathbf{r}$ typically contains intermediate solution steps followed by a final answer~\citep{wei2022chain}. Following prior work~\citep{cobbe2021training,lu2025when}, we evaluate correctness by applying a rule-based answer extractor $\mathrm{Ans}(\cdot)$ to obtain the final answer $\mathbf{a}=\mathrm{Ans}(\mathbf{r})$, then matching it against the ground-truth answer $\mathbf{a}^*$ from the dataset $\mathcal{D}=\{(\mathbf{q}_i,\mathbf{a}_i^*)\}_{i=1}^{n}$.

\vspace{0.1cm}
\subsection{Inference Modes}
Given this setup, we distinguish two inference modes of an LLM by how it allocates additional computation before emitting the response $\mathbf{r}$.

\paragraph{Thinking Mode.}
To produce logically grounded responses, recent reasoning LLMs generate long thinking traces before producing a step-by-step response containing the final solution~\citep{qwen2.5, deepseekr1, olmo3, qwen3}.
\begin{equation}
\mathbf{c} \sim p_\theta(\cdot \mid \mathbf{q}), \qquad
\mathbf{r} \sim p_\theta(\cdot \mid \mathbf{c}, \mathbf{q}).
\label{eq:internal_cot}
\end{equation}
The thinking trace $\mathbf{c}$ denotes intermediate reasoning tokens, typically learned during post-training with a reinforcement learning objective. While this mode has shown strong gains on complex tasks by allocating additional computation at inference time, it introduces significant latency overhead because the hidden chain $\mathbf{c}$ is usually much longer than the response $\mathbf{r}$ and must also be generated autoregressively.

\paragraph{Non-Thinking Mode.}
In classical settings, LLMs generate the response $\mathbf{r}$ directly, conditioned only on the query:
\begin{equation}
\mathbf{r} \sim p_\theta(\cdot \mid \mathbf{q}).
\label{eq:nothink}
\end{equation}
By sacrificing the ability to use thinking traces as a scratchpad, this mode can achieve dramatically lower latency than Thinking mode. This Non-Thinking mode can also be invoked in post-trained models by prompting with null thinking tokens~\citep{ma2025reasoning}, or by post-training with the Non-Thinking mode~\citep{qwen3}. Note that even without thinking tokens, the response tokens $\mathbf{r}$ produced in Non-Thinking mode can still include step-by-step solutions (\emph{i.e.,} chain-of-thought) before emitting the answer.
\vspace{0.3cm}
\section{\ours{}}
\label{sec:method}

We introduce \ours{}, a framework that reframes thinking as query-conditioned parameter modulation of a language model rather than explicit token generation.
We view the thinking process as an intermediate computation that steers the model’s decoding dynamics before producing the final response.
\ours{} realizes this steering through a lightweight query-specific parameter update predicted by an amortized hypernetwork, enabling direct response generation without thinking tokens while approximating the response distribution of Thinking Mode.

\vspace{0.1cm}
\subsection{Thinking Process as Parameter Updates}
We relate the Thinking and Non-Thinking modes from Section~\ref{sec:preliminaries} through the lens of distribution steering.
In Thinking mode, the chain of thinking tokens $\mathbf{c}$ is generated from each query $\mathbf{q}$ and steers the model from the direct response distribution $p_\theta(\mathbf{r}\mid\mathbf{q})$ toward the conditioned distribution $p_\theta(\mathbf{r}\mid\mathbf{q},\mathbf{c})$.
Our goal is to approximate this steering effect with a query-conditioned parameter shift $\Delta \theta(\mathbf{q})$, avoiding both generation of and conditioning on the thinking tokens $\mathbf{c}$.

Concretely, we posit the existence of a parameter update $\Delta\theta(\mathbf{q})$ such that the Non-Thinking mode decoding of the updated model matches the Thinking mode decoding:
\begin{equation}
p_{\theta+\Delta\theta(\mathbf{q})}(\mathbf{r}\mid\mathbf{q})
\;\approx\;
p_\theta(\mathbf{r}\mid\mathbf{q},\mathbf{c}).
\label{eq:steering_to_update}
\end{equation}
Here, $\Delta\theta(\mathbf{q})$ is intended to implicitly capture the effect of conditioning on $\mathbf{c}$, \emph{i.e.}, the resulting shift in the model’s activations and KV-cache.
This formulation is motivated by studies on the duality between in-context learning and fine-tuning~\citep{von2023transformers,dai2023can}.

A simple way to obtain $\Delta \theta(\mathbf{q})$ from a query $\mathbf{q}$ is to maximize the likelihood of the responses from Thinking mode under the updated model:
\begin{equation}
\Delta\theta^*(\mathbf{q})
\;=\;
\underset{\Delta\theta \in \Theta}{\mathrm{argmax}}
\underset{(\mathbf{r}, \mathbf{c}) \sim p_\theta(\cdot \mid \mathbf{q})}{\mathbb{E}} \left[
\log p_{\theta+\Delta\theta}(\mathbf{r}\mid \mathbf{q})
\right]
\label{eq:oracle_update}
\end{equation}

where $\Theta$ denotes the space of the LLM parameters.
While Eq.~\eqref{eq:oracle_update} provides a principled way to obtain $\Delta\theta^*(\mathbf{q})$, it still requires generating the thinking tokens $\mathbf{c}$ in addition to the optimization process for each query $\mathbf{q}$, which is computationally prohibitive at inference time and provides no benefit over Thinking Mode.

We therefore amortize the mapping from queries to parameter updates by training a hypernetwork $f_\psi$ that predicts $\Delta\theta$ directly from $\mathbf{q}$:
\begin{equation}
\Delta\theta(\mathbf{q}) \;=\; f_\psi(\mathbf{q}),
\label{eq:hypernet_delta}
\end{equation}
The hypernetwork can be trained end-to-end with the maximum likelihood objective in Eq.~\eqref{eq:oracle_update}.
\looseness=-10000
At inference, \ours{} applies the predicted parameter update from the trained hypernetwork to the frozen base model and directly generates the response tokens $\mathbf{r}$.
This replaces long sequential thinking-trace generation with one non-autoregressive hypernetwork pass.
The resulting method incurs a modest latency overhead over Native Non-Thinking while avoiding the substantially longer sequential decoding required by explicit thinking traces.

\vspace{0.1cm}
\subsection{Bias-Only Adaptation}
\label{subsec:bias_only}

Predicting the full weight updates $\Delta \theta$ for LLMs with a hypernetwork, however, is impractical due to the massive (\emph{e.g.,} billions) output dimensionality.
We therefore restrict adaptation to a tiny subset of parameters: \emph{bias parameters} of several selected transformer layers of the LLM backbone.
Let $\mathbf{b}\in\mathbb{R}^D$ denote the concatenation of the selected bias parameters.
\ours{} only predicts the shift in selected biases: $\Delta \mathbf{b}(\mathbf{q}) = f_\psi(\mathbf{q})$.
Then the predicted bias updates are applied to the base LLM to adapt $\theta$:
\begin{equation}
    \theta + \Delta \theta = \theta \oplus \Delta \mathbf{b}(\mathbf{q}),
    \label{eq:bias_tuning}
\end{equation}
where $\oplus$ denotes adding $\Delta\mathbf{b}(\mathbf{q})$ only to the bias subset while keeping other parameters fixed.

We choose bias parameters for two reasons.
First, the number of bias parameters is negligible compared to full weight matrices, making them easy to predict and apply with a lightweight hypernetwork.
For example, in the instantiation we show in the experiments, we reduce the parameter space of a 0.6B model to only 90K ($0.015\%$).
Second, despite this constrained subspace, bias updates have been shown to effectively steer transformer backbones across diverse domains~\citep{zaken2022bitfit,cho2024meta,beft,kim2025hyperflow}.
Other parameter subspaces, such as LoRA~\citep{hu2022lora} or Prompt Tuning~\citep{lester2021power}, are also possible in principle; we study the impact of the chosen adaptation subspace in Section~\ref{subsec:ablations}.

\begin{figure*}[t]
    \centering
    \includegraphics[width=\linewidth]{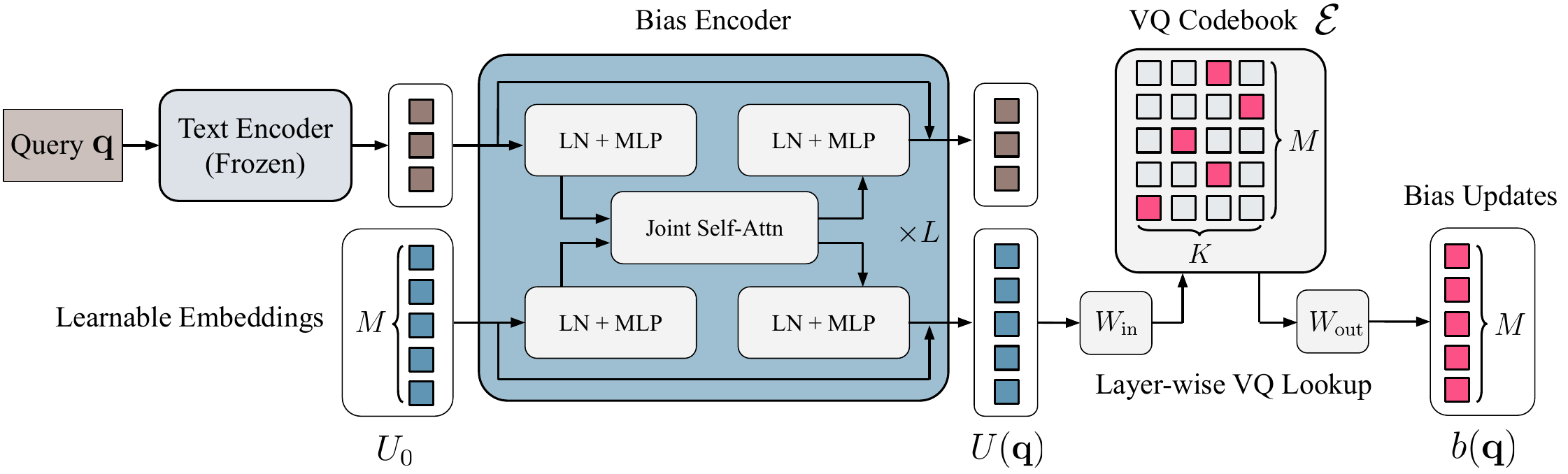}
    \caption{\textbf{\ours{} architecture.}
    A frozen text encoder embeds the query $\mathbf{q}$ into text embeddings, which are jointly encoded with learnable embeddings $U_0$ by a bias encoder.
    The output bias tokens $U(\mathbf{q})$ are decoded to final bias updates $\Delta \mathbf{b}(\mathbf{q})$ by a VQ decoder that discretizes the contextualized parameter representations using a learned codebook $\mathcal{E}$.}
    \label{fig:architecture}
    \vspace{0.2cm}
\end{figure*}

\vspace{0.1cm}
\subsection{Hypernetwork Architecture}
\label{subsec:architecture}

The hypernetwork $f_\psi$ consists of three components (see Figure~\ref{fig:architecture}): a \emph{Text Encoder} that embeds the query, a \emph{Bias Encoder} that contextualizes a set of learnable parameter tokens with the query representation, and a \emph{VQ Decoder} that discretizes the resulting representations and projects them into the target bias-update space.

\paragraph{Text Encoder.}
The first stage extracts a contextual representation of the input query $\mathbf{q}$.
We employ a frozen pretrained transformer (\emph{e.g.,} a copy of the base LLM) to produce a sequence of text tokens:
\begin{equation}
H_{\text{text}} = \mathrm{Encoder}_{\text{text}}(\mathbf{q}) \in \mathbb{R}^{L_q \times d_{\text{enc}}},
\end{equation}
where $L_q$ is the query length.
These embeddings serve as the query context for subsequent parameter generation.
By using a pretrained language model as the text encoder, we obtain robust representations that capture the semantic structure of the query relevant for predicting bias updates.
Since the query length is relatively small and we only apply a single forward pass of the encoder, the cost of text encoding is substantially lower than autoregressively generating a long chain of thinking tokens, and thus does not conflict with our goal of efficient reasoning.

\paragraph{Bias Encoder.}
The hypernetwork must output a \emph{structured} bias update, consisting of multiple bias vectors for each layer.
To explicitly represent these update slots, we introduce $M$ learnable \emph{bias tokens}
$U_0=[u_1;\dots;u_M]\in\mathbb{R}^{M\times d_{\text{hyper}}}$,
where $M$ is the number of layers whose biases are updated and each token $u_j$ is assigned to a layer $\ell_j$.
Then we encode the bias tokens using an MM-DiT~\citep{esser2024scaling} architecture, which jointly encodes the text tokens $H_{\text{text}}$ and the parameter tokens $U_0$ by considering them as separate modalities, with modality-specific LayerNorm/MLP and joint self-attention.
This enables the model to contextualize query information into the parameters while capturing the dependencies among layer-wise parameters.
We use the output parameter tokens as the contextualized representation
$U(\mathbf{q})\in\mathbb{R}^{M\times d_{\text{hyper}}}$,
which will be decoded into the final bias update.

\paragraph{VQ Decoder.}
To regularize the update space and encourage reuse of common update patterns, we introduce a discrete bottleneck via vector quantization (VQ) with layer-wise learnable codebooks $\mathcal{E}^j=\{e_1^j,\dots,e_K^j\}\subset\mathbb{R}^{d_z}$ to decode the continuous outputs of the bias encoder.
For each token $U_j(\mathbf{q})$, we project it to a latent $z_j$, quantize it to the nearest code within the corresponding codebook $\mathcal{E}^j$, and decode it into the actual bias update $\Delta \mathbf{b}^{(\ell_j)}$ for the layer.
Concatenating $\Delta\mathbf{b}^{(\ell_j)}(\mathbf{q})$ over $j=1,\dots,M$ yields the full bias update $\Delta\mathbf{b}(\mathbf{q})\in\mathbb{R}^{D}$.
During training, we use a straight-through estimator to backpropagate through the quantization step, while learning the codebook entries jointly with the decoder.
The discrete bottleneck acts as an inductive bias by constraining updates to be composed from a finite set of learned prototypes shared by queries.
This reduces the effective degrees of freedom of instance-specific adaptation and makes the predicted update behave more like a reusable reasoning primitive than an unconstrained per-example perturbation.
We zero-initialize the VQ decoder's output projection $(W_{\mathrm{out}},b_{\mathrm{out}})$, ensuring $\Delta\mathbf{b}(\mathbf{q})\equiv0$ and an initial match to the frozen base model.

\vspace{0.1cm}
\subsection{Training Objective}
\label{subsec:training_objective}

\begin{figure*}[t]
    \centering
    \includegraphics[trim={0cm 1cm 0.2cm 0cm},clip,width=\linewidth]{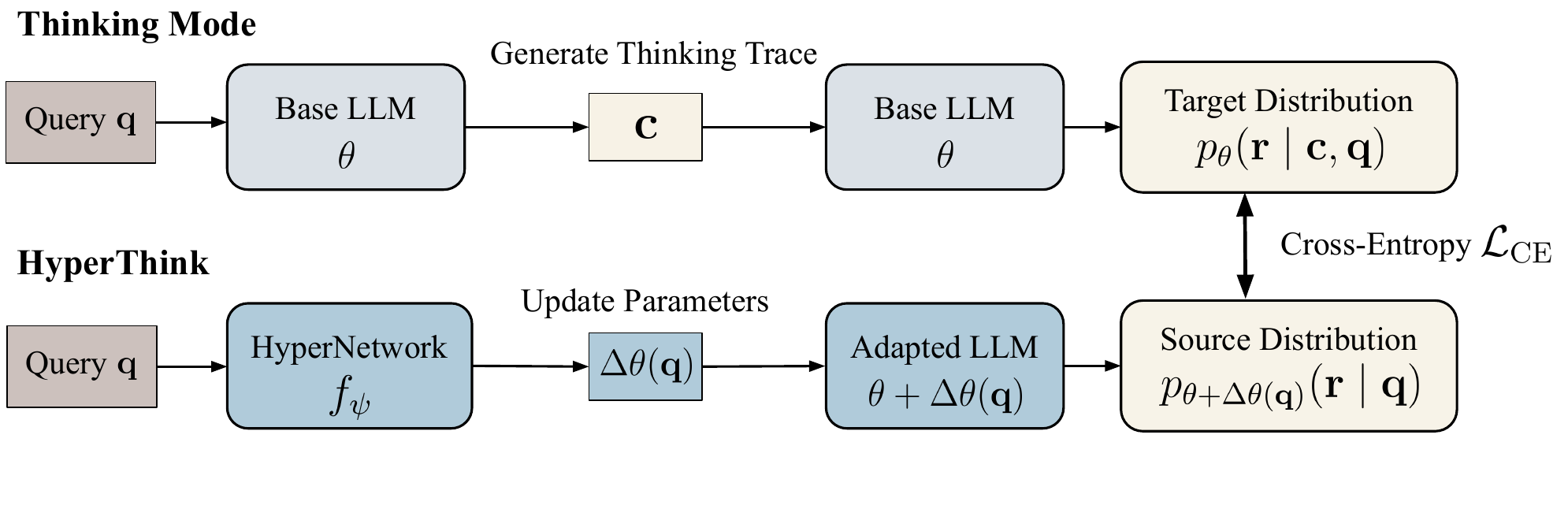}
    \caption{Maximum likelihood objective for \ours{}.
    \ours{} predicts query-specific bias parameters and injects them into the base LLM, and is trained to steer the Non-Thinking response distribution into that of Thinking mode.}
    \label{fig:context_distillation}
    \vspace{0.2cm}
\end{figure*}

During training, the base parameters $\theta=(\theta_{\text{w}},\mathbf{b})$ are kept frozen; we update only the hypernetwork parameters $\psi$ and the VQ codebook $\mathcal{E}$ end-to-end.
Our objective consists of (i) a maximum-likelihood term that trains \ours{} to reproduce teacher-generated responses without internal thinking tokens, (ii) a vector-quantization objective that stabilizes the discrete bottleneck, and (iii) a usage regularizer that prevents code collapse in the VQ decoder.

\paragraph{Maximum Likelihood Objective.} To encourage the hypernetwork output $f_\psi(\mathbf{q})$ to steer the response distribution in the Non-Thinking mode toward that of the Thinking mode, we employ an end-to-end maximum likelihood (cross-entropy) objective using correctness-filtered, teacher-generated response tokens from Thinking Mode:
\begin{equation}
\mathcal{L}_{\mathrm{CE}}(\psi)
=
\underset{\substack{\mathbf{q}\sim\mathcal{D} \\ \mathbf{r}, \mathbf{c} \sim p_\theta(\cdot\mid \mathbf{q})}}{\mathbb{E}}
\left[
\sum_{t=1}^{|\mathbf{r}|}
-\log p_{\theta + \Delta\theta(\mathbf{q})}\bigl(r_t \mid \mathbf{r}_{<t}, \mathbf{q}\bigr)
\right],
\label{eq:HyperThink_ce}
\end{equation}
where the parameter updates $\Delta\theta(\mathbf{q})$ are produced by the hypernetwork $f_\psi$ (Eq.~\eqref{eq:bias_tuning}).
This objective steers the adapted LLM to \emph{skip} the thinking tokens $\mathbf{c}$ while producing the logically grounded responses $\mathbf{r} \sim p_\theta(\cdot\mid\mathbf{c}, \mathbf{q})$.

\paragraph{Vector Quantization Objective.} To learn discrete reasoning prototypes, we employ the standard VQ-VAE objective~\citep{van2017neural}.
We use a straight-through estimator to backpropagate through the non-differentiable code selection and add the auxiliary loss
\begin{equation}
\mathcal{L}_{\mathrm{VQ}}(\psi)
=
\underset{\mathbf{q} \sim \mathcal{D}}{\mathbb{E}}
\Bigg[
\sum_{j=1}^M \Big(
\| \mathrm{sg}[z_j(\mathbf{q})] - e_{k_j} \|_2^2
+
\beta \| z_j(\mathbf{q}) - \mathrm{sg}[e_{k_j}] \|_2^2
\Big)
\Bigg],
\label{eq:vq_loss}
\end{equation}
where $z_j(\mathbf{q})$ is the projected latent for parameter token $j$, $e_{k_j}$ is the selected code, $\mathrm{sg}[\cdot]$ denotes stop-gradient, and $\beta \in \mathbb{R}$ is a hyperparameter.

\paragraph{Usage Regularizer.}
Optimizing the VQ objective alone can lead to \emph{code collapse}, where most latents are assigned to a few codes.
To encourage diverse usage, we regularize the batch-wise code distribution.
For each bias group $j$, we aggregate soft VQ assignments over a mini-batch of size $B$ to obtain an empirical usage distribution $p_j(k)$ over the $K$ codes, and encourage it to be close to uniform by minimizing
\begin{equation}
\mathcal{L}_{\mathrm{Use}}
=
\mathbb{E}_{\mathbf{q}_1,\dots,\mathbf{q}_B \sim \mathcal{D}}
\left[
\sum_{j=1}^M
D_{\mathrm{KL}}\!\left(p_j(k) \,\middle\|\, \mathcal{U}(k)\right)
\right]
\end{equation}
where $\mathcal{U}(k)$ is the uniform distribution over $K$ codes. We use a straight-through estimator to backpropagate through $p_j(k)$.
This increases code-usage entropy and mitigates collapse.

\paragraph{Total Objective.}
The total objective combines all terms:
\begin{equation}
\mathcal{L}_{\mathrm{total}} = \mathcal{L}_{\mathrm{CE}} + \lambda_{\mathrm{VQ}} \mathcal{L}_{\mathrm{VQ}} + \lambda_{\mathrm{Use}} \mathcal{L}_{\mathrm{Use}},
\end{equation}
where $\lambda_{\mathrm{VQ}}, \lambda_{\mathrm{Use}} \in \mathbb{R}$ are hyperparameters.

\vspace{0.3cm}
\section{Experiments}
\label{sec:experiments}

In this section, we empirically evaluate \ours{} on various LLM reasoning tasks under two distinct experimental setups: math reasoning and general reasoning.

\subsection{Experimental Setup}
\label{subsec:exp_setup}

\paragraph{Base LLMs.} 
For the math reasoning experiments, we use Qwen3-0.6B~\citep{qwen3} and SmolLM3-3B~\citep{bakouch2025smollm3} as base models.
For general reasoning experiments, we use SmolLM3-3B and Olmo-3-7B-Think~\citep{olmo3} as base models.
All models are extensively post-trained for Thinking and Non-Thinking inference, and Qwen3 supports token-budget control, making them exceptionally strong baselines for \ours{}.
We follow their standard decoding settings, with a temperature of $0.6$ and nucleus sampling with $p=0.95$.
For each query, we draw five samples and report the average accuracy and Pass@5 as performance metrics.
To fairly assess the inference cost of our hypernetwork-based method and prompting baselines, we also report FLOPs averaged over the samples.

\paragraph{Datasets.}
For the mathematical reasoning experiments, we train Qwen3-0.6B on GSM8K~\citep{cobbe2021training} and SmolLM3-3B on DeepScaleR~\citep{deepscaler2025}.
Then we evaluate both models on the standard GSM8K test split as well as MATH-500~\citep{hendrycks2021measuring} as the out-of-domain benchmark.
For the general reasoning experiments, we augment the training data with CodeForces-CoTs~\citep{penedo2025codeforces}, LogiQA~\citep{liu2020logiqa}, OpenBookQA~\citep{OpenBookQA2018}, and QASC~\citep{qasc}, also covering coding, logical reasoning, and general question-answering tasks.
We then evaluate the model on AIME~\citep{aime_2024,aime_2025}, LiveCodeBench~\citep{jain2025livecodebench}, BIG-Bench Hard~\citep{suzgun2023challenging}, and CommonsenseQA~\citep{talmor2019commonsenseqa}, which assess Olympiad-level mathematics, code generation, multi-step reasoning, and commonsense question answering, respectively.

\paragraph{Baselines.}
We compare \ours{} against five baselines: \textbf{Thinking Mode}, which generates an explicit reasoning trace before producing the final response; \textbf{Budget-Controlled Thinking Mode}, which constrains Thinking-Mode generation to a token budget comparable to that of \ours{}; \textbf{Native Non-Thinking Mode}, which directly produces a response using the model's native non-thinking configuration; \textbf{System~2 Distillation}~\citep{yu2024distilling}, which uses the same training objective as \ours{} but directly fine-tunes the base LLM rather than predicting query-conditioned parameter updates; and \textbf{TokenSkip}~\citep{tokenskip}, which learns to selectively omit less important tokens for controllable compression of reasoning traces.
Full details are provided in Appendix~\ref{appendix:baselines}.

\paragraph{Implementation Details.}
For the hypernetwork text encoder, we use a pretrained backbone and keep it frozen during training. The hypernetwork consists of a lightweight MM-DiT-based bias encoder and a VQ decoder.
We adapt biases in the later transformer blocks of the base LLM, targeting a small subset of projection modules; the exact architectures and parameter counts for each backbone are given in Appendix~\ref{appendix:implementation_details}.
Unless otherwise noted, all other hypernetwork and training settings are shared across backbones. We train \ours{} and the baselines with AdamW~\citep{loshchilov2019decoupled}; full optimization hyperparameters and compute details are deferred to Appendix~\ref{appendix:implementation_details}.

\vspace{0.1cm}
\subsection{Main Results}
\label{subsec:main_results}

Table~\ref{tab:main_results} summarizes the Qwen3-0.6B results on the mathematical-reasoning benchmarks.
We observe that \ours{} generally improves over the Non-Thinking and distillation baselines with only a small increase in FLOPs.
In particular, at similar compute budgets, \ours{} yields significant improvements over Budget-Controlled Thinking Mode on both benchmarks.
We also observe that System~2 Distillation and TokenSkip underperform \ours{} on the out-of-distribution MATH-500 benchmark.
This suggests weaker out-of-distribution generalization with less-constrained fine-tuning, whereas VQ-constrained bias adaptation preserves stronger transfer to MATH-500.
On the other hand, \ours{} successfully generalizes to out-of-distribution math problems to perform reasoning without thinking tokens.
In the SmolLM3-3B model, as summarized in Table~\ref{tab:main_results_smollm3b}, the gap between \ours{} and baselines even widens.
\ours{} generally outperforms the low-budget baselines with much lower FLOPs, indicating that \ours{} scales well with model size and has strong potential for practical applications.

\begin{figure}[!t]
    \centering
    \includegraphics[width=0.95\linewidth]{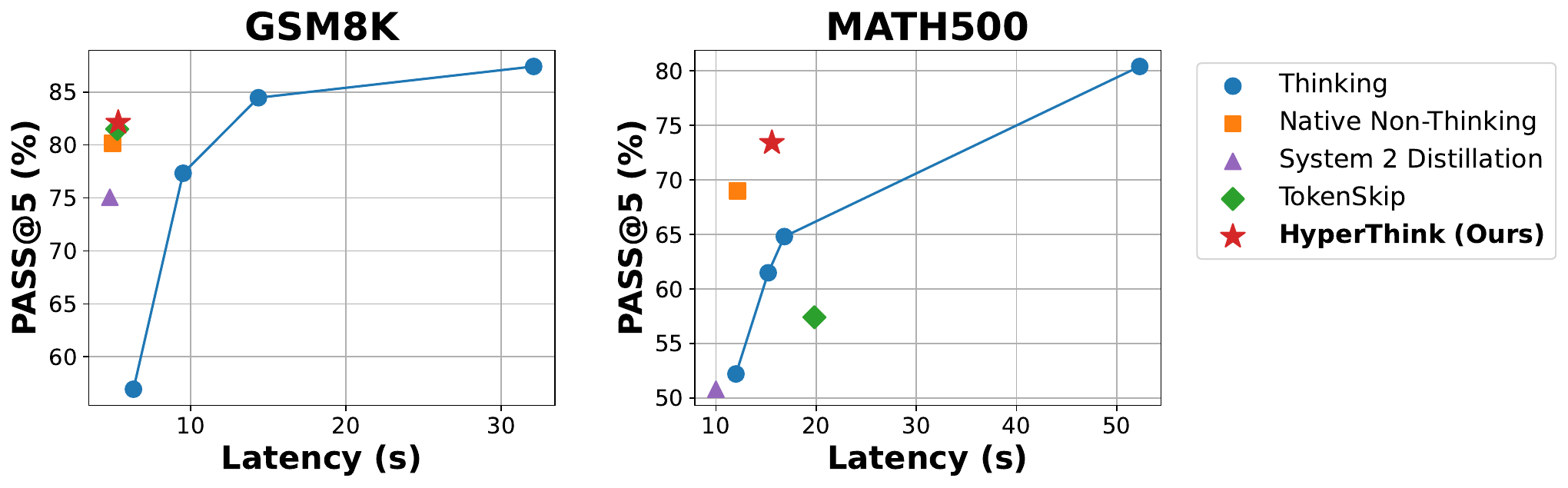}
    \caption{Average Pass@5 versus latency of Qwen3-0.6B on math reasoning tasks.}
    \vspace{0.2cm}
    \label{fig:flops}
    \vspace{0.2cm}
\end{figure}

To clearly investigate the cost-performance trade-off between baselines, we plot Pass@5 against the end-to-end answering latency measured on a single NVIDIA H200 GPU in Figure~\ref{fig:flops}.
Due to the lightweight computation of the hypernetwork, which is a single forward pass on the problem text, the latency of \ours{} is largely determined by the number of generated tokens.
Since \ours{} learns to directly generate response tokens, it incurs latency comparable to Native Non-Thinking.
Therefore, \ours{} achieves a significantly better accuracy--latency trade-off than Budget-Controlled Thinking Mode and System~2 Distillation.
Moreover, \ours{} still provides a performance boost over the Native Non-Thinking Mode with a slight increase in latency, which is notable because the base Qwen3 model is already post-trained for the Non-Thinking mode with an extensive corpus over multiple training stages.

\begin{table}
    \begin{center}
        \renewcommand{\arraystretch}{1.3}
        \renewcommand{\aboverulesep}{0pt}
        \renewcommand{\belowrulesep}{0pt}
        
        \setlength{\tabcolsep}{4pt}
        \scriptsize
        \begin{tabularx}{\textwidth}{@{}>{\raggedright\arraybackslash}Xcccccc@{}}
            \toprule
            
            \multirow{2}{*}{\textbf{Reasoning Mode}}
            & \multicolumn{3}{c}{\textbf{GSM8K-Test}}
            & \multicolumn{3}{c}{\textbf{MATH-500}}
            \\
            
            \cmidrule(lr){2-4} \cmidrule(lr){5-7}
            & Acc (\%)
            & Pass@5 (\%)
            & FLOPs (G)
            & Acc (\%)
            & Pass@5 (\%)
            & FLOPs (G)
            \\
            
            \midrule
            
            Thinking Mode~\citep{qwen3}
            & 73.81 & 87.41 & 2503.71 & 52.64 & 80.40 & 4535.70 \\
            
            \midrule

            Budget-Controlled Thinking Mode~\citep{qwen3}
            & 36.15 & 56.94 & 502.36 & 43.76 & 61.48 & 1198.27 \\

            Native Non-Thinking Mode~\citep{qwen3}
            & 58.82 & 80.14 & 407.66 & 46.84 & 69.00 & 962.35 \\

            System~2 Distillation~\citep{yu2024distilling}
            & 61.91 & 75.06 & 458.21 & 31.40 & 50.80 & 838.52 \\

            TokenSkip~\citep{tokenskip}
            & 61.64 & 81.50 & 437.18 & 37.12 & 57.40 & 1562.24 \\

            \textbf{\ours{} (Ours)}
            & 61.06 & 82.11 & 477.28 & 48.76 & 73.40 & 1150.83 \\         
        
            \bottomrule
        \end{tabularx}
    \end{center}
    \caption{
    Main results on GSM8K-Test and MATH-500 with Qwen3-0.6B backbone.
    We report average accuracy, Pass@5, and FLOPs for inference over 5 samples for all methods.
    }
    \label{tab:main_results}
    \vspace{0.2cm}
\end{table}

\begin{table}[t]
    \vspace{0.05in}
    \centering
    \renewcommand{\arraystretch}{1.3}
    \renewcommand{\aboverulesep}{0pt}
    \renewcommand{\belowrulesep}{0pt}
    \setlength{\tabcolsep}{4pt}
    \scriptsize

    \begin{tabularx}{\textwidth}{@{}>{\raggedright\arraybackslash}Xcccccc@{}}
        \toprule
        
        \multirow{2}{*}{\textbf{Reasoning Mode}}
        & \multicolumn{3}{c}{\textbf{GSM8K-Test}}
        & \multicolumn{3}{c}{\textbf{MATH-500}} \\
        
        \cmidrule(lr){2-4} \cmidrule(lr){5-7}
        & Acc (\%)
        & Pass@5 (\%)
        & FLOPs (G)
        & Acc (\%)
        & Pass@5 (\%)
        & FLOPs (G)
        \\
        
        \midrule

        Thinking Mode~\citep{bakouch2025smollm3}
        & 92.27 & 96.51 & 8340.31 & 88.56 & 93.60 & 28460.97 \\

        \midrule

        Budget-Controlled Thinking Mode~\citep{qwen3}
        & 62.23 & 83.02 & 3748.67 & 46.80 & 63.20 & 5645.65 \\

        Native Non-Thinking Mode~\citep{bakouch2025smollm3}
        & 74.81 & 95.15 & 4900.09 & 70.20 & 84.20 & 7143.64 \\
        
        System~2 Distillation~\citep{yu2024distilling}
        & 74.72 & 89.99 & 3400.70 & 62.00 & 79.60 & 5510.84 \\

        TokenSkip~\citep{tokenskip}
        & 81.08 & 93.10 & 3015.82 & 54.56 & 74.00 & 10991.38 \\

        \textbf{\ours{} (Ours)}
        & 84.75 & 95.15 & 3303.77 & 67.36 & 81.20 & 5900.99 \\
        
        \bottomrule
    \end{tabularx}
    \vspace{0.2cm}
    \caption{
    Main results on GSM8K-Test and MATH-500 with SmolLM3-3B backbone.
    We report average accuracy, Pass@5, and FLOPs for inference over 5 samples for all methods.
    }
    \label{tab:main_results_smollm3b}
    \vspace{0.2cm}
\end{table}

\begin{figure}[!t]
    \centering
    \includegraphics[width=0.95\linewidth]{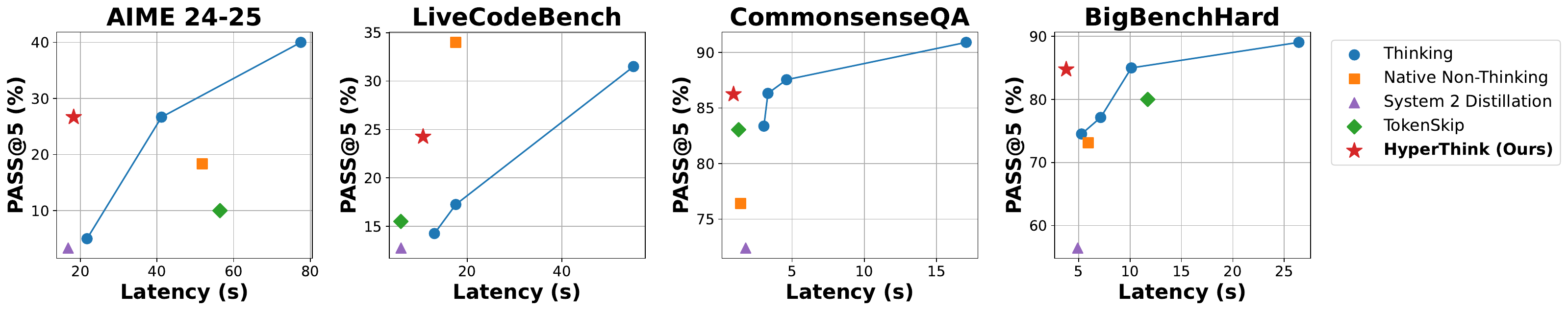}
    \vspace{0.2cm}
    \caption{Average Pass@5 versus latency of SmolLM3-3B on general reasoning tasks.}
    \label{fig:smol_results}
    \vspace{0.4cm}
\end{figure}

\begin{figure}[!t]
    \centering
    \includegraphics[width=0.95\linewidth]{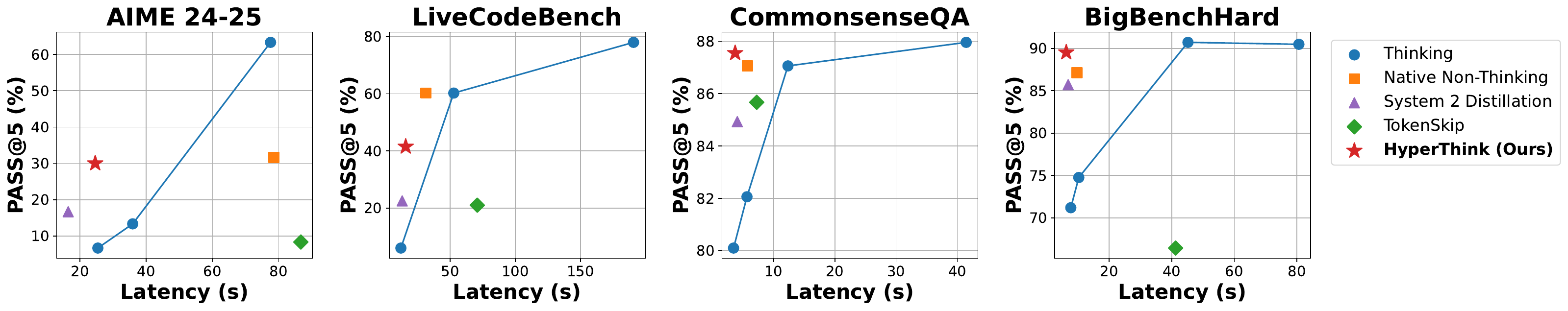}
    \vspace{0.2cm}
    \caption{Average Pass@5 versus latency of Olmo-3-7B-Think on general reasoning tasks.}
    \label{fig:olmo_results}
    \vspace{0.4cm}
\end{figure}

Figures~\ref{fig:smol_results} and~\ref{fig:olmo_results} report results for SmolLM3-3B and Olmo-3-7B-Think, respectively, with both models trained on the same mixed-domain corpus.
Across CommonsenseQA and BIG-Bench Hard, \ours{} generally improves over Native Non-Thinking in the low-latency regime, showing that query-conditioned parameter updates can support reasoning beyond mathematics when the corresponding domains are covered during training.
In particular, on Olmo-3-7B-Think, \ours{} achieves performance comparable to the evaluated Thinking-mode operating points on these QA benchmarks while requiring only approximately $7$--$9\%$ of their answering latency.
The harder AIME and LiveCodeBench evaluations clarify the boundary of this trade-off: high-budget Thinking remains substantially stronger on search-intensive Olympiad problems, while Native Non-Thinking can remain competitive or stronger on executable code generation.
Thus, \ours{} does not uniformly dominate all baselines, but provides a favorable operating point near the Non-Thinking latency regime across larger backbones and broader reasoning domains.

We provide the complete results for the general-reasoning experiments in Appendix~\ref{appendix:complete_reasoning_results}.

\begin{table}[!t]
    \begin{center}
        \renewcommand{\arraystretch}{1.3}
        \renewcommand{\aboverulesep}{0pt}
        \renewcommand{\belowrulesep}{0pt}
        \scriptsize
        \begin{tabular}{ccccccccc}
    
            \toprule
            
            \multirow{2}{*}{\textbf{Bias Only}}
            & \multirow{2}{*}{\textbf{TTA}}
            & \multirow{2}{*}{\textbf{VQ Bottleneck}}
            & \multicolumn{3}{c}{\textbf{GSM8K-Test}}
            & \multicolumn{3}{c}{\textbf{MATH-500}}
            \\
            
            \cmidrule(lr){4-6} \cmidrule(lr){7-9}

            & &
            & Acc (\%)
            & Pass@5 (\%)
            & FLOPs (G)
            & Acc (\%)
            & Pass@5 (\%)
            & FLOPs (G)
            \\
            
            \midrule

            \xmark & \xmark & \xmark
            & 61.91 & 75.06 & 458.21 & 31.40 & 50.80 & 838.52 \\

            \cmark & \xmark & \xmark
            & 60.88 & 82.11 & 440.72 & 45.56 & 66.20 & 1065.50 \\

            \cmark & \cmark & \xmark
            & 59.74 & 80.14 & 425.39 & 43.32 & 66.90 & 1046.11 \\

            \cmark & \cmark & \cmark
            & 61.06 & 82.11 & 477.28 & 48.76 & 73.40 & 1150.83 \\

            \bottomrule
        \end{tabular}
    \end{center}
    \vspace{0.2cm}
    \caption{
    Component-wise ablation study on \ours{} architecture.
    }
    \label{tab:componentwise_ablation}
    \vspace{0.2cm}
\end{table}

\vspace{0.1cm}
\subsection{Ablation Studies}
\label{subsec:ablations}

In this section, we provide further studies on the main design choices in \ours{} and the learned VQ codebook, using Qwen3-0.6B as the base model on the math experiment.
We leave additional ablation studies about the choice of target bias parameters and additional analysis of the VQ bottleneck in Appendix~\ref{appendix:additional_ablations}.

\paragraph{Component-wise Analysis.}
To demonstrate the effectiveness of each architectural component of \ours{}, we report the performance of its ablated variants in Table~\ref{tab:componentwise_ablation}.
The first row corresponds to the System~2 Distillation baseline, which uses the same distillation objective and training data as \ours{} but fully fine-tunes the LLM.
The second row restricts this baseline to optimizing only bias parameters, which can be viewed as global biases shared across queries.
The large improvement over the first row highlights the robustness of bias parameters for steering a pretrained LLM.
The third row is \ours{} without the VQ bottleneck, directly projecting the continuous outputs of the Bias Encoder to bias updates, and the last row is the full \ours{}.
The results show that simply introducing query-specific test-time adaptation with the Text/Bias Encoders is insufficient to generalize the query-to-parameter mapping (Eq.~\eqref{eq:oracle_update}).
With the VQ decoder, \ours{} achieves much better out-of-distribution generalization on MATH-500, indicating that the discrete bottleneck acts as a strong regularizer against overfitting.

\begin{wraptable}[7]{r}{0.5\textwidth}
    \vspace{-0.075in}
    \scriptsize
    \setlength{\tabcolsep}{2pt} 
    \begin{center}
        \renewcommand{\arraystretch}{1.2}
        \renewcommand{\aboverulesep}{0pt}
        \renewcommand{\belowrulesep}{0pt}
        \begin{tabular}{cccccc}
            \toprule
            \multirow{2}{*}{\textbf{\shortstack[c]{Adaptable\\Parameters}}}
            & \multicolumn{2}{c}{\textbf{GSM8K-Test}}
            & \multicolumn{2}{c}{\textbf{MATH-500}} \\
            \cmidrule(lr){2-3} \cmidrule(lr){4-5}
            & {\scriptsize Acc (\%)} &  {\scriptsize FLOPs (G)}
            & {\scriptsize Acc (\%)} &  {\scriptsize FLOPs (G)} \\
            \midrule
            LoRA
            & 60.71 & 503.44 & 46.92 & 1130.62 \\
            Prompt Tuning
            & 60.18 & 506.68 & 43.52 & 879.76 \\
            \textbf{Bias (Ours)}
            & 61.06 & 477.28 & 48.76 & 1150.83 \\
            \bottomrule
        \end{tabular}
    \end{center}
    \caption{
    Ablation of adaptable parameters.
    }
    \label{tab:param_ablation}
\end{wraptable}

\paragraph{Effect of the Adaptable Parameters.}
As discussed in Section~\ref{subsec:bias_only}, \ours{} can predict updates in parameter subspaces other than biases.
To justify our choice, we compare variants of \ours{} using two common parameter-efficient fine-tuning approaches: LoRA~\citep{hu2022lora} and Prompt Tuning~\citep{lester2021power}.
We use rank $8$ for LoRA and $32$ prompt tokens for Prompt Tuning.
Table~\ref{tab:param_ablation} summarizes the results.
We observe that both alternative parameterizations underperform bias adaptation, which supports the flexibility and robustness of bias-only tuning.

\begin{wrapfigure}[20]{r}{0.5\textwidth}
    \vspace{-0.2cm}
    \centering
    \includegraphics[width=\linewidth]{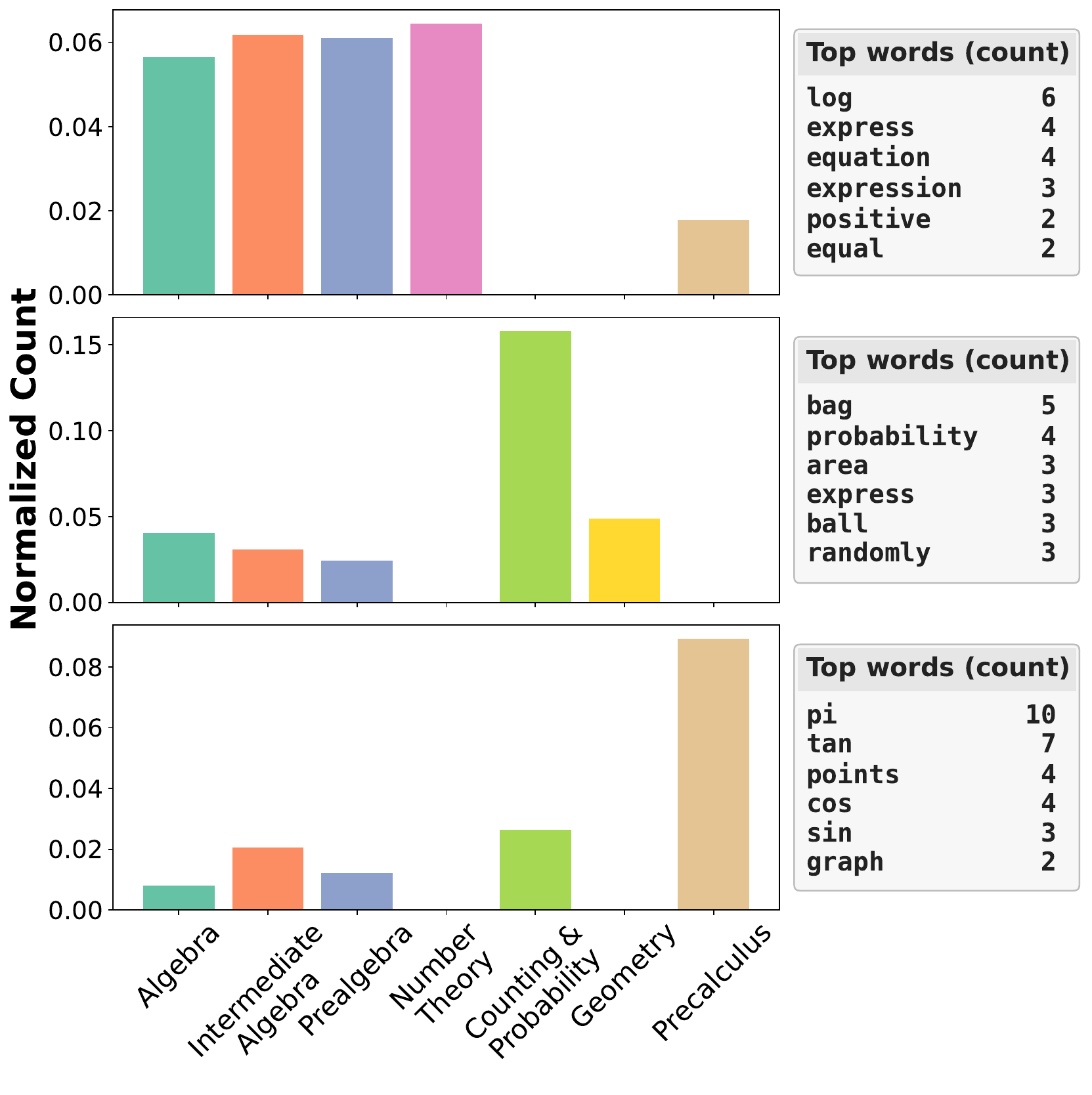}
    \caption{Subject distribution and top keywords per VQ code.}
    \label{fig:vq_analysis}
\end{wrapfigure}

\paragraph{Analysis of the Learned VQ Codebook.}
To investigate the learned structure in the VQ codebook, we analyze whether similar problems are assigned the same VQ code.
Figure~\ref{fig:vq_analysis} reports the distribution of subjects in MATH-500 for three representative VQ codes, along with the most frequent keywords in the problems assigned to each code.
We observe that each code primarily covers related subjects.
For example, the first code is mainly used for Algebra, Intermediate Algebra, Prealgebra, and Number Theory, and it is dominated by algebraic keywords such as ``log'' and ``equation''.
The second and third codes are used mainly for Probability and Precalculus, respectively, and are dominated by subject-related keywords.
Overall, the alignment between subject-level assignments and code-specific lexical patterns suggests that individual VQ codes capture meaningful semantic structure in the math problem space.

\vspace{0.3cm}
\section{Related Work}
\label{sec:related_work}

\paragraph{Efficient Reasoning with Large Language Models.} Chain-of-Thought (CoT) prompting unlocks strong reasoning in LLMs~\citep{wei2022chain, kojima2022large} but introduces substantial inference latency. To address this, \citet{ning2023skeleton} parallelize reasoning steps via hierarchical prompting, \citet{tokenskip} prune redundant reasoning tokens, and \citet{thinkprune} constrain LLMs to produce shorter traces during post-training. A complementary line of work distills reasoning from strong teachers into student models via thinking traces and step-by-step supervision~\citep{magister2023teaching, ho2023large, sprague2025skillfactory, yu2024distilling}. These methods show that reasoning can be compiled into efficient ``System~1''-like behavior, but the result is a single static set of student weights. \ours{} shares the goal of distilling System~2 into System~1 reasoning~\citep{yu2024distilling}, but instead of training a student to directly output final answers, it amortizes thinking traces into dynamic, query-conditional parameter updates applied to a frozen base model.

\paragraph{Hypernetworks for Parameter Generation.} Hypernetworks~\citep{ha2016hypernetworks} generate target-network weights conditioned on an input and have been adopted for parameter-efficient multi-task adaptation in NLP~\citep{mahabadi2021parameter, charakorn2025text}, where a shared hypernetwork produces task-specific adapters~\citep{hu2022lora} reused across all problems of a given task. More recently, instance-level adaptation~\citep{ifpruning, zhyper} predicts parameters conditioned on each input example; HyperFlow~\citep{kim2025hyperflow} emulates gradient-based few-shot fine-tuning by predicting task-conditional drifts without explicit backpropagation. Our work shares this philosophy of ``immediate query-specific adaptation,'' but targets LLM reasoning rather than few-shot visual classification.

We present additional related work on test-time adaptation in Appendix~\ref{appendix:related_work}.

\vspace{0.3cm}
\section{Conclusion}
\label{sec:conclusion}

\looseness=-10000
We introduced \ours{}, a text-to-parameter framework that amortizes long thinking traces into query-conditional parameter updates.
Rather than generating explicit thinking tokens at test time, \ours{} predicts a small, structured parameter shift for each query, enabling the base model to operate in the efficient \emph{non-thinking} mode while emulating the reasoning ability of the \emph{thinking} mode.
At the core of our approach is a bias-predicting hypernetwork equipped with a vector-quantized bottleneck, which constrains per-query adaptations to a finite set of reusable reasoning prototypes.
Empirically, \ours{} improves the low-latency region of the accuracy--latency trade-off on mathematical and general reasoning tasks, with its strongest gains in the near-non-thinking regime.

\clearpage

\section*{Acknowledgements}
This work was supported by the Institute of Information \& Communications Technology Planning \& Evaluation (IITP) grants funded by the Ministry of Science and ICT (MSIT) of the Republic of Korea (No. RS-2024-00469482, No. RS-2024-00509279, No. RS-2019-II190075). This work was also supported by the InnoCORE program of the MSIT (AI Meta-Scientist, N10260110), and in part by the “HPC support” project funded by the Korea government.

\section*{Ethics Statement}
This paper proposes a method for amortizing long-form reasoning traces into query-conditioned parameter updates, improving the efficiency of LLM inference on mathematical reasoning tasks. While this work is motivated by reducing computational cost in settings such as math problem solving, efficiency improvements to LLM reasoning could, in principle, lower the barrier to deploying reasoning models for harmful purposes. This concern is not unique to our approach but applies broadly to research that improves the cost-performance tradeoff of LLM inference.

\section*{Reproducibility Statement}
We describe our experimental setup in Section~\ref{sec:experiments}, including details on base LLMs, training datasets, and evaluation benchmarks. Full architectural details of the hypernetwork, including the text encoder, bias encoder, and VQ decoder configurations, are provided in Section~\ref{sec:method} and Appendix~\ref{appendix:implementation_details}. We specify training hyperparameters (optimizer, learning rate, batch size, loss weights) in Appendix~\ref{appendix:implementation_details}. All base models used in our experiments are publicly available. We plan to release both training and evaluation code upon publication.

\bibliography{colm2026_conference}
\bibliographystyle{colm2026_conference}

\clearpage
\appendix
\clearpage
\section*{\LARGE Appendix}

\vspace{0.3cm}
\section{Related Work}\label{appendix:related_work}

In this section, we provide further discussion of related work on test-time adaptation, latent reasoning, and context distillation methods.

\vspace{0.1cm}
\paragraph{Test-Time Adaptation.}
A complementary direction to test-time scaling via longer reasoning traces is to adapt models at inference time by updating either parameters or internal representations.
Test-time training has been explored across various domains such as image classification~\citep{originalttt, tttmaskedauto}, language modeling~\citep{hardt2024test}, and abstract reasoning~\citep{akyurek2025the, lu2025context}.
Relatedly, diffusion models employ classifier guidance, using objectives that repeatedly shape intermediate states during generation~\citep{classifierguidance, glide, lu2024procreate}.
In this view, the core question is how to trade additional inference-time compute for improved task performance or controllability.
\ours{} shares this perspective by introducing an explicit per-query weight adaptation step.
However, instead of performing gradient-based optimization, it amortizes the LLM's reasoning process into a query-conditioned update predicted by a hypernetwork's single forward pass, enabling fast per-query test-time adaptation.

\paragraph{Latent Reasoning.}
Latent-reasoning methods seek to move intermediate computation from natural-language tokens into continuous hidden states.
Coconut~\citep{hao2024training}, for example, treats the LLM's last hidden state as a continuous thought and feeds it back as the next input embedding, allowing multiple intermediate reasoning steps without decoding each step into language.
This is conceptually related to \ours{} because both reduce reliance on explicit textual thinking traces.
Their inference mechanisms are different, however.
Coconut modifies and trains the base LLM to consume a sequence of recurrently generated latent states, and therefore retains sequential intermediate computation in the latent space.
In contrast, \ours{} leaves the base LLM frozen and predicts a single temporary parameter update from the query through one non-autoregressive hypernetwork pass.
The adapted model then decodes a short response without executing a recurrent latent trajectory.
We therefore regard Coconut as a complementary latent-reasoning approach rather than a directly comparable frozen-backbone baseline.

\paragraph{Context and Adapter Distillation.}
Context distillation trains a context-free student to reproduce the behavior of a teacher that receives privileged context, such as instructions, demonstrations, scratchpads, or documents~\citep{snell2022distillingcontext}.
Recent work stores the effect of such context in parameter-efficient modules.
Deep Context Distillation~\citep{caccia2025deepcontext}, for example, trains document-specific LoRA modules to match the hidden states and output distributions of a teacher that has access to the document.
Related hypernetwork methods generate adapters from task descriptions or textual context~\citep{charakorn2025text,zhyper}, and Doc-to-LoRA~\citep{charakorn2026doctolora} amortizes document context distillation into a single adapter-generation pass.

\ours{} shares the counterfactual structure of context distillation: its teacher has access to privileged context that is withheld from the student at inference.
In our case, however, the privileged context is the base model's own query-specific thinking trace rather than a persistent external document or task description.
Moreover, \ours{} does not produce one static student or one reusable document-level adapter.
It keeps the base LLM frozen and maps each query to a temporary VQ-regularized bias update whose purpose is to approximate the downstream effect of that query's reasoning computation.
Thus, context/adapter distillation and \ours{} share the principle of translating transient contextual computation into parameter space, but differ in the source, granularity, and lifetime of the distilled information.

\vspace{0.3cm}
\section{Further Details on Baselines}
\label{appendix:baselines}

In this section, we describe the baselines used throughout the experiments in greater detail.

\begin{itemize}[left=4pt,topsep=0pt,itemsep=0pt]
    \item \textbf{Thinking Mode} runs the base LLM using its standard Thinking Mode configuration to generate thinking tokens encapsulated by special tokens (\emph{i.e.}, {\footnotesize \texttt{<think>}} $\cdots$ {\footnotesize \texttt{</think>}}) before producing the final response.
    This setting serves as an upper bound that exploits an unconstrained token budget.

    \item \textbf{Budget-Controlled Thinking Mode} runs Thinking mode while constraining the token budget to a level comparable to that of \ours{}, to precisely assess the accuracy--computation trade-off.
    Specifically, we follow the procedure of the Qwen3 paper~\citep{qwen3} where we stop once generation reaches the specified budget and force the model to directly produce an answer by appending an instruction prompt.

    \item \textbf{Native Non-Thinking Mode} runs the base LLM by prompting with an empty thinking trace (\emph{i.e.}, {\footnotesize \texttt{<think>} \texttt{</think>}} with no content in between) and directly produces the response conditioned only on the question.
    Since the base models are post-trained to directly produce responses under the non-thinking prompt, this serves as the strongest baseline without any additional adaptation to the base LLM.

    \item \textbf{System~2 Distillation} uses the same maximum likelihood objective as \ours{} to skip the thinking tokens, but directly fine-tunes the base LLM for all problems in the training dataset.
    This approach is a minor modification of \citet{yu2024distilling}, which distills high-quality responses after System~2 reasoning into System~1 generation, whereas we apply this idea to distill the effect of thinking traces into direct generation of step-by-step solutions.

    \item \textbf{TokenSkip}~\citep{tokenskip} first compresses the teacher-generated CoT traces using the LLMLingua-2~\citep{pan2024llmlingua} token-importance estimator.
    Conditioned on the query and a target compression ratio, the model is trained with the standard autoregressive cross-entropy objective to generate the compressed CoT followed by the unchanged final response.
    We use the same training queries and teacher traces as \ours{} and follow the original LoRA SFT configuration, keeping the base LLM frozen and optimizing only rank-$8$ LoRA adapters with scaling factor $\alpha=16$.
\end{itemize}

\vspace{0.3cm}
\section{Further Details on Implementation}
\label{appendix:implementation_details}

In this section, we further describe the implementation details of \ours{} and the baselines used throughout the experiments.

For the hypernetwork text encoder, we use a pretrained LLM backbone and keep it frozen during training: Qwen3-Embedding-0.6B~\citep{qwen3} for the Qwen3-0.6B experiments, and the SmolLM3-3B and Olmo-3-7B-Think models themselves for their respective experiments.
The bias encoder consists of three MM-DiT~\citep{esser2024scaling} blocks with hidden size $128$, and the VQ decoder employs a codebook with $K=256$ entries and latent dimension $512$.
For Qwen3-0.6B, we adapt the biases in the last eight transformer blocks, where each block contains biases for six projection modules, \texttt{q\_proj}, \texttt{v\_proj}, \texttt{o\_proj}, \texttt{up\_proj}, \texttt{gate\_proj}, and \texttt{down\_proj}, totaling $90{,}122$ parameters.
For SmolLM3-3B and Olmo-3-7B-Think, we instead adapt the biases in the last half of the transformer blocks, using the same set of projection modules, yielding totals of $516{,}096$ and $614{,}400$ parameters, respectively.
Unless otherwise noted, all other hypernetwork and training settings are shared across backbones.
We train \ours{} and the baselines with the AdamW optimizer~\citep{loshchilov2019decoupled}, a learning rate of $1\times10^{-4}$, and a batch size of $16$.
For \ours{}, we use $\lambda_{\mathrm{VQ}}=100$ and $\lambda_{\mathrm{Use}}=0.1$.
We run all training experiments on 2 NVIDIA H200 GPUs and evaluation on a single H200 GPU.

\vspace{0.3cm}
\section{Complete Results on General Reasoning Experiments}
\label{appendix:complete_reasoning_results}
In this section, we provide complete quantitative results of the general reasoning experiments with SmolLM3-3B and Olmo-3-7B-Think base models.
Tables~\ref{tab:smol_full_results} and~\ref{tab:olmo_full_results} report the results of each model with average accuracy, Pass@5, and FLOPs on the four general reasoning benchmarks: AIME, LiveCodeBench, CommonsenseQA, and BIG-Bench Hard.

\begin{table}
    \begin{center}
        \renewcommand{\arraystretch}{1.3}
        \renewcommand{\aboverulesep}{0pt}
        \renewcommand{\belowrulesep}{0pt}
        
        \setlength{\tabcolsep}{4pt}
        \scriptsize
        \begin{tabularx}{\textwidth}{@{}>{\raggedright\arraybackslash}Xcccccc@{}}
            \toprule
            
            \multirow{2}{*}{\textbf{Reasoning Mode}}
            & \multicolumn{3}{c}{\textbf{AIME}}
            & \multicolumn{3}{c}{\textbf{LiveCodeBench}}
            \\
            
            \cmidrule(lr){2-4} \cmidrule(lr){5-7}
            & Acc (\%)
            & Pass@5 (\%)
            & FLOPs (G)
            & Acc (\%)
            & Pass@5 (\%)
            & FLOPs (G)
            \\
            
            \midrule
            
            Thinking Mode~\citep{bakouch2025smollm3}
            & 41.00 & 66.67 & 125796.78 & 42.80 & 63.50 & 136267.93 \\
            
            \midrule

            Budget-Controlled Thinking Mode~\citep{qwen3}
            & 4.33 & 5.00 & 9541.25 & 9.05 & 14.25 & 7361.08 \\

            Native Non-Thinking Mode~\citep{bakouch2025smollm3}
            & 9.00 & 18.33 & 20458.29 & 19.75 & 34.00 & 10572.33 \\

            System~2 Distillation~\citep{yu2024distilling}
            & 1.00 & 3.33 & 8443.08 & 7.80 & 12.75 & 6640.46 \\

            TokenSkip~\citep{tokenskip}
            & 3.33 & 10.00 & 21778.68 & 9.90 & 15.50 & 23672.06 \\

            \textbf{\ours{} (Ours)}
            & 8.33 & 26.67 & 8607.56 & 13.55 & 24.25 & 11163.28 \\
        
            \bottomrule
        \end{tabularx}
        \vspace{0.1cm}
        
        \begin{tabularx}{\textwidth}{@{}>{\raggedright\arraybackslash}Xcccccc@{}}
            \toprule
            
            \multirow{2}{*}{\textbf{Reasoning Mode}}
            & \multicolumn{3}{c}{\textbf{CommonsenseQA}}
            & \multicolumn{3}{c}{\textbf{BIG-Bench Hard}}
            \\
            
            \cmidrule(lr){2-4} \cmidrule(lr){5-7}
            & Acc (\%)
            & Pass@5 (\%)
            & FLOPs (G)
            & Acc (\%)
            & Pass@5 (\%)
            & FLOPs (G)
            \\
            
            \midrule
            
            Thinking Mode~\citep{bakouch2025smollm3}
            & 76.81 & 90.91 & 7248.80 & 71.43 & 89.05 & 11179.57 \\
            
            \midrule

            Budget-Controlled Thinking Mode~\citep{qwen3}
            & 72.29 & 83.37 & 2736.48 & 48.57 & 74.52 & 4137.30 \\

            Native Non-Thinking Mode~\citep{bakouch2025smollm3}
            & 49.58 & 76.41 & 2290.83 & 44.48 & 73.10 & 4360.30 \\

            System~2 Distillation~\citep{yu2024distilling}
            & 59.26 & 72.40 & 2409.94 & 37.33 & 56.43 & 4099.58 \\

            TokenSkip~\citep{tokenskip}
            & 73.33 & 83.05 & 2234.44 & 54.10 & 80.00 & 6257.30 \\

            \textbf{\ours{} (Ours)}
            & 71.30 & 86.24 & 2083.52 & 56.90 & 84.76 & 3709.57 \\
        
            \bottomrule
        \end{tabularx}
    \end{center}
    \vspace{0.2cm}
    \caption{
    Main results on four general reasoning benchmarks with SmolLM3-3B backbone.
    We report average accuracy, Pass@5, and FLOPs for inference over 5 samples for all methods.
    }
    \label{tab:smol_full_results}
    \vspace{0.2cm}
\end{table}

\begin{table}
    \begin{center}
        \renewcommand{\arraystretch}{1.3}
        \renewcommand{\aboverulesep}{0pt}
        \renewcommand{\belowrulesep}{0pt}
        
        \setlength{\tabcolsep}{4pt}
        \scriptsize
        \begin{tabularx}{\textwidth}{@{}>{\raggedright\arraybackslash}Xcccccc@{}}
            \toprule
            
            \multirow{2}{*}{\textbf{Reasoning Mode}}
            & \multicolumn{3}{c}{\textbf{AIME}}
            & \multicolumn{3}{c}{\textbf{LiveCodeBench}}
            \\
            
            \cmidrule(lr){2-4} \cmidrule(lr){5-7}
            & Acc (\%)
            & Pass@5 (\%)
            & FLOPs (G)
            & Acc (\%)
            & Pass@5 (\%)
            & FLOPs (G)
            \\
            
            \midrule
            
            Thinking Mode~\citep{olmo3}
            & 69.00 & 83.33 & 301968.24 & 76.90 & 93.25 & 200170.05 \\
            
            \midrule

            Budget-Controlled Thinking Mode~\citep{qwen3}
            & 3.33 & 6.67 & 19103.66 & 2.60 & 6.00 & 15693.76 \\

            Native Non-Thinking Mode~\citep{olmo3}
            & 18.00 & 31.67 & 52997.12 & 40.35 & 60.25 & 28264.01 \\

            System~2 Distillation~\citep{yu2024distilling}
            & 6.00 & 16.67 & 12918.68 & 14.10 & 22.50 & 16312.32 \\

            TokenSkip~\citep{tokenskip}
            & 2.33 & 8.33 & 60068.92 & 11.65 & 21.00 & 56019.72 \\

            \textbf{\ours{} (Ours)}
            & 13.00 & 30.00 & 17322.58 & 24.75 & 41.50 & 17328.44 \\
        
            \bottomrule
        \end{tabularx}
        \vspace{0.1cm}
        
        \begin{tabularx}{\textwidth}{@{}>{\raggedright\arraybackslash}Xcccccc@{}}
            \toprule
            
            \multirow{2}{*}{\textbf{Reasoning Mode}}
            & \multicolumn{3}{c}{\textbf{CommonsenseQA}}
            & \multicolumn{3}{c}{\textbf{BIG-Bench Hard}}
            \\
            
            \cmidrule(lr){2-4} \cmidrule(lr){5-7}
            & Acc (\%)
            & Pass@5 (\%)
            & FLOPs (G)
            & Acc (\%)
            & Pass@5 (\%)
            & FLOPs (G)
            \\
            
            \midrule
            
            Thinking Mode~\citep{olmo3}
            & 78.74 & 87.96 & 27570.45 & 80.52 & 90.48 & 56189.67 \\
            
            \midrule

            Budget-Controlled Thinking Mode~\citep{qwen3}
            & 71.73 & 80.10 & 3922.13 & 50.43 & 71.19 & 7929.79 \\

            Native Non-Thinking Mode~\citep{olmo3}
            & 73.01 & 87.06 & 5359.36 & 69.48 & 87.14 & 9538.29 \\

            System~2 Distillation~\citep{yu2024distilling}
            & 64.44 & 84.93 & 4242.58 & 53.00 & 85.71 & 7718.65 \\

            TokenSkip~\citep{tokenskip}
            & 68.86 & 85.67 & 6255.19 & 36.95 & 66.43 & 29609.53 \\

            \textbf{\ours{} (Ours)}
            & 69.42 & 87.55 & 4019.39 & 63.24 & 89.52 & 6888.85 \\
        
            \bottomrule
        \end{tabularx}
    \end{center}
    \vspace{0.2cm}
    \caption{
    Main results on four general reasoning benchmarks with Olmo-3-7B-Think backbone.
    We report average accuracy, Pass@5, and FLOPs for inference over 5 samples for all methods.
    }
    \label{tab:olmo_full_results}
    \vspace{0.2cm}
\end{table}

\vspace{0.3cm}
\section{Additional Ablation Studies}
\label{appendix:additional_ablations}

In this section, we provide additional ablation studies of the target bias parameters and an analysis of the VQ bottleneck.

\paragraph{Choice of Target Bias Parameters.}
We use the same target module types across all backbones: the biases of \texttt{q\_proj}, \texttt{v\_proj}, \texttt{o\_proj}, \texttt{up\_proj}, \texttt{gate\_proj}, and \texttt{down\_proj}.
We exclude \texttt{k\_proj} because prior analyses suggest that its bias is largely redundant with the query-projection bias in dot-product attention~\citep{zaken2022bitfit,namazifar2023role}.
The only backbone-specific choice is the adapted layer range.
We focus on later transformer blocks, which are more directly involved in high-level reasoning and answer formation, while leaving earlier lexical and syntactic representations unchanged.
Accordingly, we update the last 8 blocks of Qwen3-0.6B and the last half of SmolLM3-3B and Olmo-3-7B-Think.
Table~\ref{tab:layer_sensitivity} confirms that this choice is robust: the last-8 setting performs best on MATH-500 for Qwen3-0.6B, while the last-half setting performs best on both GSM8K-Test and MATH-500 for SmolLM3-3B.

\begin{table}[t]
    \begin{center}
        \renewcommand{\arraystretch}{1.3}
        \renewcommand{\aboverulesep}{0pt}
        \renewcommand{\belowrulesep}{0pt}
        \small
        \begin{tabular}{lcccccccc}
            \toprule
            \multirow{2}{*}{\textbf{Bias Layers}}
            & \multicolumn{4}{c}{\textbf{Qwen3-0.6B}}
            & \multicolumn{4}{c}{\textbf{SmolLM3-3B}} \\
            \cmidrule(lr){2-5} \cmidrule(lr){6-9}
            & \multicolumn{2}{c}{\textbf{GSM8K-Test}} & \multicolumn{2}{c}{\textbf{MATH-500}}
            & \multicolumn{2}{c}{\textbf{GSM8K-Test}} & \multicolumn{2}{c}{\textbf{MATH-500}} \\
            & \textbf{Acc} & \textbf{Pass@5} & \textbf{Acc} & \textbf{Pass@5} & \textbf{Acc} & \textbf{Pass@5} & \textbf{Acc} & \textbf{Pass@5} \\
            \midrule
            All         & 63.44 & 83.62 & 40.84 & 61.20 & 82.24 & 92.27 & 59.36 & 75.40 \\
            First 8     & 62.52 & 81.73 & 43.84 & 65.00 & 82.27 & 92.72 & 63.52 & 77.40 \\
            First half  & 63.62 & 83.32 & 41.36 & 62.80 & 81.85 & 92.49 & 61.04 & 76.40 \\
            Last half   & 60.59 & 81.58 & 42.52 & 64.80 & 84.75 & 95.15 & 67.36 & 81.20 \\
            Last 8      & 61.06 & 82.11 & 48.76 & 73.40 & 83.46 & 93.63 & 63.16 & 78.00 \\
            \bottomrule
        \end{tabular}
    \end{center}
    \vspace{0.2cm}
    \caption{Sensitivity analysis on the target bias layers. We use the later-layer window as the default: last 8 blocks for Qwen3-0.6B and last half of blocks for SmolLM3-3B.}
    \label{tab:layer_sensitivity}
    \vspace{0.2cm}
\end{table}

\paragraph{Additional Analysis of the VQ Bottleneck.}
The main ablation in Section~\ref{subsec:ablations} shows that the VQ bottleneck improves
held-out performance, but it does not by itself distinguish improved
optimization from improved generalization. We therefore additionally evaluate
the continuous and VQ-regularized variants on the GSM8K training set.
The two variants perform nearly identically on the GSM8K training set, while
the difference becomes larger on held-out data.
VQ substantially improves accuracy on GSM8K-Test and MATH-500, so the bottleneck does not materially improve fitting of the training examples.
Instead, the gain emerges primarily on unseen queries and is largest under the GSM8K-to-MATH-500 distribution shift.
This supports the interpretation that VQ regularizes the query-to-parameter mapping by forcing related queries to reuse a finite collection of update prototypes.
Together with the semantic codebook analysis in Figure~\ref{fig:vq_analysis}, these results indicate that the discrete bottleneck improves transfer rather than merely increasing training-set fit.

\begin{table}
    \begin{center}
        \renewcommand{\arraystretch}{1.3}
        \renewcommand{\aboverulesep}{0pt}
        \renewcommand{\belowrulesep}{0pt}
        \small
        \begin{tabular}{lcccccc}
        \toprule
        & \multicolumn{2}{c}{\textbf{GSM8K-Train}}
        & \multicolumn{2}{c}{\textbf{GSM8K-Test}}
        & \multicolumn{2}{c}{\textbf{MATH-500}} \\
        \cmidrule(lr){2-3}
        \cmidrule(lr){4-5}
        \cmidrule(lr){6-7}
        \textbf{Variant}
        & \textbf{Acc. (\%)} & \textbf{Pass@5 (\%)}
        & \textbf{Acc. (\%)} & \textbf{Pass@5 (\%)}
        & \textbf{Acc. (\%)} & \textbf{Pass@5 (\%)} \\
        \midrule
        \ours{} w/o VQ
        & 67.44 & 86.83
        & 59.74 & 80.14
        & 43.32 & 66.90 \\
        \ours{}
        & 67.53 & 86.67
        & 61.06 & 82.11
        & 48.76 & 73.40 \\
        \bottomrule
        \end{tabular}
    \end{center}
    \vspace{0.2cm}
    \caption{Train--test diagnostic for the VQ bottleneck using Qwen3-0.6B
    trained on GSM8K. Both variants are evaluated using five sampled responses
    per query.}
    \label{tab:vq_train_test}
    \vspace{0.2cm}
\end{table}

\vspace{0.3cm}
\section{Example Generation for Each Inference Mode}
\label{sec:generation_examples}

In this section, we provide generation examples for each inference mode compared in the experiments: Thinking Mode, Native Non-Thinking Mode, System~2 Distillation, and \ours{}.

\vspace{0.2cm}
\subsection{GSM8K Examples}

\begin{InteractionBox}{Thinking Mode}

\FieldHeader{Problem}
\begin{LLMBlock}
<|im_start|>user
Jill gets paid $20 per hour to teach and $30 to be a cheerleading coach. If she works 50 weeks a year, 35 hours a week as a teacher and 15 hours a week as a coach, what's her annual salary?
\end{LLMBlock}

\FieldHeader{Prompt}
\begin{LLMBlock}
Please reason step by step, and put your final answer within \boxed{}.<|im_end|>
<|im_start|>assistant
\end{LLMBlock}

\FieldHeader{Thinking Trace}
\begin{LLMBlock}
<think>
Okay, let's see. Jill's annual salary is calculated by multiplying her hourly wage by the number of hours she works each week, and then multiplying that by the number of weeks she works in a year. 

First, I need to confirm the numbers given. The problem says she gets paid $20 per hour for teaching and $30 for cheerleading coaching. She works 50 weeks a year, 35 hours a week as a teacher, and 15 hours a week as a coach. 

So, let me break it down. For the teaching part: 20 dollars per hour times 35 hours a week. Then, for the cheerleading coach part: 30 dollars per hour times 15 hours a week. Then, multiply all those by the number of weeks, which is 50. 

Wait, let me check if I need to do anything else. The problem says ``annual salary,'' so I think that's just adding the weekly salaries together. 

Let me calculate each part step by step. 

First, teacher's hourly rate: 20 dollars/hour * 35 hours/week. Let me compute that. 

20 * 35. Hmm, 20 times 30 is 600, and 20 times 5 is 100, so total is 600 + 100 = 700 dollars per week for teaching. 

Then, cheerleading coach's hourly rate: 30 dollars/hour * 15 hours/week. Let me do that. 

30 * 15. 30 times 10 is 300, and 30 times 5 is 150, so total is 300 + 150 = 450 dollars per week for coaching. 

Now, total weekly salary is 700 (teaching) + 450 (coach) = 1150 dollars per week. 

Then, multiply by the number of weeks, which is 50. 

So, 1150 * 50. Let me calculate that. 

Well, 1000 * 50 is 50,000, and 150 * 50 is 7,500. Adding those together: 50,000 + 7,500 = 57,500. 

So, her annual salary would be $57,500. 

Wait, let me double-check my calculations to make sure I didn't make any errors. 

First, teacher's weekly rate: 20 * 35. Yes, 20*35. Let me do it another way: 35*20. 35*2*10=70*10=700. Correct. 

Coach's weekly rate: 30*15. 30*10=300, 30*5=150, so 300+150=450. Correct. 

Total weekly: 700+450=1150. Correct. 

Then 1150*50. Let me verify that multiplication again. 

50 weeks * 1150 dollars per week. 

Alternatively, 50*1000=50,000 and 50*150=7,500. So total is 57,500. Yes, that's correct. 

I think that's all. So the answer should be $57,500. 

**Final Answer**
Jill's annual salary is \boxed{57500}.
</think>
\end{LLMBlock}

\FieldHeader{Response}
\begin{LLMBlock}
To determine Jill's annual salary, we need to calculate her weekly earnings and then multiply that by the number of weeks in a year.

---

### **Step 1: Weekly Earnings**

- **Teaching:**  
  Jill earns $20 per hour for teaching.  
  She works **35 hours a week**.  
  **Earnings:**  
  $$
  20 \times 35 = 700 \text{ dollars per week}
  $$

- **Cheering Coach:**  
  Jill earns $30 per hour for coaching.  
  She works **15 hours a week**.  
  **Earnings:**  
  $$
  30 \times 15 = 450 \text{ dollars per week}
  $$

---

### **Step 2: Total Weekly Earnings**

Adding both parts of her income:  
$$
700 + 450 = 1150 \text{ dollars per week}
$$

---

### **Step 3: Annual Earnings**

Jill works **50 weeks a year**.  
$$
1150 \times 50 = 57500 \text{ dollars}
$$

---

### **Final Answer**

$$
\boxed{57500}
$$
\end{LLMBlock}
\end{InteractionBox}

\begin{InteractionBox}{Native Non-Thinking Mode}

\FieldHeader{Problem}
\begin{LLMBlock}
<|im_start|>user
Jill gets paid $20 per hour to teach and $30 to be a cheerleading coach. If she works 50 weeks a year, 35 hours a week as a teacher and 15 hours a week as a coach, what's her annual salary?
\end{LLMBlock}

\FieldHeader{Prompt}
\begin{LLMBlock}
Please reason step by step, and put your final answer within \boxed{}.<|im_end|>
<|im_start|>assistant
<think>

</think>
\end{LLMBlock}

\FieldHeader{Response}
\begin{LLMBlock}
Response:
To find Jill's **annual salary**, we need to calculate her **hourly earnings** and multiply by the **number of hours she works per year**.

### Step 1: Calculate the hourly earnings

- **Teacher pay**: $20 per hour  
- **Coach pay**: $30 per hour  

So, her **hourly earnings** are:  
$$
20 + 30 = 50 \text{ dollars per hour}
$$

### Step 2: Calculate total hours per year

- She works **50 weeks** a year  
- She works **35 hours** a week as a teacher  
- She works **15 hours** a week as a coach  

Total hours per year:  
$$
50 \text{ weeks} \times (35 \text{ hours/week} + 15 \text{ hours/week}) = 50 \times 50 = 2500 \text{ hours}
$$

### Step 3: Calculate annual salary

$$
2500 \text{ hours} \times 50 \text{ dollars/hour} = \boxed{125000} \text{ dollars}
$$

### Final Answer:

$$
\boxed{125000}
$$
\end{LLMBlock}
\end{InteractionBox}

\begin{InteractionBox}{System~2 Distillation}

\FieldHeader{Prompt}
\begin{LLMBlock}
Please reason step by step, and put your final answer within \boxed{{}}.
\end{LLMBlock}

\FieldHeader{Problem}
\begin{LLMBlock}
Jill gets paid $20 per hour to teach and $30 to be a cheerleading coach. If she works
50 weeks a year, 35 hours a week as a teacher and 15 hours a week as a coach,
what's her annual salary?
\end{LLMBlock}

\FieldHeader{Response}
\begin{LLMBlock}
Response:
To determine Jill's annual salary, we need to calculate her total earnings from both her teacher and coach roles over the course of the year.

---

### Step 1: Calculate Weekly Earnings

- **Teacher's Hourly Rate:** $20 per hour  
  - **Hours per week:** 35 hours  
  - **Teacher's weekly earnings:**  
    $$
    35 \times 20 = 700
    $$

- **Coach's Hourly Rate:** $30 per hour  
  - **Hours per week:** 15 hours  
  - **Coach's weekly earnings:**  
    $$
    15 \times 30 = 450
    $$

---

### Step 2: Calculate Annual Earnings

- **Teacher's annual earnings:**  
  $$
  700 \times 50 = 35,000
  $$

- **Coach's annual earnings:**  
  $$
  450 \times 50 = 22,500
  $$

---

### Step 3: Total Annual Salary

$$
35,000 + 22,500 = 57,500
$$

---

### Final Answer

$$
\boxed{57500}
$$
\end{LLMBlock}
\end{InteractionBox}

\begin{InteractionBox}{\ours{}}

\FieldHeader{Problem}
\begin{LLMBlock}
<|im_start|>user
Jill gets paid $20 per hour to teach and $30 to be a cheerleading coach. If she works 50 weeks a year, 35 hours a week as a teacher and 15 hours a week as a coach, what's her annual salary?
\end{LLMBlock}

\FieldHeader{Prompt}
\begin{LLMBlock}
Please reason step by step, and put your final answer within \boxed{}.<|im_end|>
<|im_start|>assistant
\end{LLMBlock}

\FieldHeader{Response}
\begin{LLMBlock}
Response:
To determine Jill's annual salary, we begin by calculating her hourly wage and then multiplying it by the number of hours she works in a year. 

Jill earns $20 per hour from teaching and $30 to be a cheerleading coach. 

Her total weekly earnings are calculated by multiplying her hourly wage by the number of hours she works each week. 

Jill works 35 hours a week as a teacher and 15 hours as a coach. 

Therefore, her weekly earnings are:
$$
35 \text{ hours} \times 20 \text{ dollars/hour} + 15 \text{ hours} \times 30 \text{ dollars/hour}
$$

Calculating this:
$$
35 \times 20 = 700
$$
$$
15 \times 30 = 450
$$

Adding these together gives her weekly earnings:
$$
700 + 450 = 1150
$$

Since she works 50 weeks a year, her annual earnings are:
$$
1150 \text{ dollars/week} \times 50 \text{ weeks/year} = 57,500
$$

Thus, Jill's annual salary is:
$$
\boxed{57500}
$$
\end{LLMBlock}
\end{InteractionBox}

\vspace{0.2cm}
\subsection{MATH-500 Examples}

\begin{InteractionBox}{Thinking Mode}

\FieldHeader{Problem}
\begin{LLMBlock}
<|im_start|>user
For some real numbers $a$ and $b$, the equation \[
8x^3 + 4ax^2 + 2bx + a = 0
\]has three distinct positive roots. If the sum of the base-2 logarithms of the roots is 5, what is the value of $a$?
\end{LLMBlock}

\FieldHeader{Prompt}
\begin{LLMBlock}
Please reason step by step, and put your final answer within \boxed{}.<|im_end|>
<|im_start|>assistant
\end{LLMBlock}

\FieldHeader{Thinking Trace}
\begin{LLMBlock}
<think>
Okay, so I need to solve this problem where there's a cubic equation: 8x^3 + 4a x^2 + 2b x + a = 0, and it has three distinct positive roots. Also, the sum of the base-2 logarithms of the roots is 5. I need to find the value of a. Alright, let me break this down step by step.

First, since it's a cubic equation with three distinct positive roots, I can use Vieta's formulas to relate the coefficients to the roots. Vieta's formulas for a cubic equation x^3 + px^2 + qx + r = 0 give me:

Sum of roots: -p
Sum of products two at a time: q
Product of roots: -r

But my equation isn't in standard form. Let me rewrite it properly. The given equation is 8x^3 + 4a x^2 + 2b x + a = 0. Let me divide each term by 8 to make the leading coefficient 1:

x^3 + (4a/8)x^2 + (2b/8)x + a/8 = 0

Simplify:

x^3 + (a/2)x^2 + (b/4)x + a/8 = 0

So, in standard form, it's x^3 + p x^2 + q x + r = 0, where p = a/2, q = b/4, r = a/8.

Therefore, applying Vieta's formulas:

1. Sum of roots: -p = -a/2
2. Sum of products two at a time: q = b/4
3. Product of roots: -r = -a/8

But Vieta's formulas also relate the roots to the coefficients. However, the problem says that the sum of the base-2 logarithms of the roots is 5. Let me recall that if a number has a logarithm base 2 of some value, say log_2(k), then k = 2^5 = 32. Since the roots are positive, their logarithms will be defined.

Let me denote the three roots as r1, r2, r3. Then:

log_2(r1) + log_2(r2) + log_2(r3) = 5.

Using logarithm properties, this sum is equal to log_2(r1 r2 r3) = 5. Therefore, r1 r2 r3 = 2^5 = 32.

So, the product of the roots is 32. But from Vieta's formula, the product of the roots is -r. Wait, let me check again.

Original equation: x^3 + (a/2)x^2 + (b/4)x + a/8 = 0. Let me write Vieta's formulas again:

Sum of roots: - (a/2) = r1 + r2 + r3.

Sum of products two at a time: (b/4) = (r1 r2 + r1 r3 + r2 r3).

Product of roots: - (a/8) = r1 r2 r3.

So, the product of the roots is -a/8. Therefore, r1 r2 r3 = -a/8. But earlier, I concluded that r1 r2 r3 = 32. Therefore, -a/8 = 32 => a = -256. Wait, but hold on, the problem says that the roots are positive. If a is negative, then the product would be negative? But the product of three positive numbers is positive. Therefore, there's a contradiction here. So, maybe my reasoning is wrong?

Wait, hold on. Let me check again. If the product of the roots is -a/8, and the roots are positive, then the product is positive. Therefore, -a/8 must be positive. Therefore, -a/8 > 0 => a < 0. So, a is negative. Therefore, the value of a is -256? But let me check again.

Wait, from Vieta's formula, product of roots is -a/8. But if the roots are positive, then their product is positive. Therefore, -a/8 must be positive. Therefore, a must be negative. So, solving -a/8 = 32 => a = -256. So, that would be the answer? But let me check if there's something wrong here.

But let me check again. The sum of logs is 5. So log_2(r1) + log_2(r2) + log_2(r3) = log_2(r1 r2 r3) = 5. Therefore, r1 r2 r3 = 32. But according to Vieta, product is -a/8. Therefore, -a/8 = 32 => a = -256. But then why does the problem mention that the roots are positive? Because if a is negative, then the coefficients of the equation might have some negative signs. Let me check the original equation again.

Original equation: 8x^3 + 4a x^2 + 2b x + a = 0. So, if a is -256, then the equation becomes 8x^3 + 4*(-256)x^2 + 2b x + (-256) = 0. Let me compute that:

8x^3 - 1024x^2 + 2b x - 256 = 0. So, the leading coefficient is 8, which is positive. The coefficient of x^2 is -1024, which is negative. The coefficient of x is 2b, which is positive if b is positive, and negative if b is negative. The constant term is -256, which is negative. So, the roots are positive? But if the product of the roots is 32, which is positive, and the product of the roots is -a/8, which would be positive if a is negative. So, that's okay. But why does the problem specify that the roots are positive? Maybe because if a is negative, then the equation might have some roots that are positive and others negative? But the problem says three distinct positive roots. So, maybe there's something else here?

Wait, but according to Vieta, product is 32. So, if the product is 32, then the roots multiply to 32. But if all roots are positive, then their product is positive. So, that's okay. So, maybe the answer is a = -256? But let me check again if my Vieta's formulas were correctly applied.

Original equation: 8x^3 + 4a x^2 + 2b x + a = 0. Let me write Vieta's formulas again:

Sum of roots: r1 + r2 + r3 = - (coefficient of x^2)/ (coefficient of x^3) = - (4a)/8 = -a/2.

Sum of products two at a time: r1 r2 + r1 r3 + r2 r3 = (coefficient of x)/ (coefficient of x^3) = (2b)/8 = b/4.

Product of roots: r1 r2 r3 = - (constant term)/ (coefficient of x^3) = -a/8.

Yes, that's correct. Therefore, product is -a/8. Since the roots are positive, product is positive. Therefore, -a/8 > 0 => a < 0. Therefore, a = -256. But let me check if there's another way to get this answer.

Alternatively, maybe I made a mistake in interpreting the sum of logs. Let me check again.

Sum of logs is log_2(r1) + log_2(r2) + log_2(r3) = log_2(r1 r2 r3) = 5. Therefore, r1 r2 r3 = 32. Correct. And Vieta gives r1 r2 r3 = -a/8. Therefore, solving -a/8 = 32 => a = -256. So, that seems straightforward.

But let me check if there's a possibility that the product is negative? But the roots are positive, so their product is positive. Therefore, -a/8 must be positive. Therefore, a must be negative. Therefore, a = -256. Therefore, the answer is \boxed{-256}?

Wait, but let me check with actual numbers. Suppose a = -256. Then, the equation is 8x^3 - 1024x^2 + 2b x - 256 = 0. Let me assume that the product of the roots is 32. Let me suppose that the roots are r1, r2, r3. Then, r1 r2 r3 = 32. Also, from Vieta, sum of roots is -a/2 = -(-256)/2 = 128. So, r1 + r2 + r3 = 128. Also, sum of products is b/4. But since we don't know b, maybe we can find b? But the problem doesn't give us more information. However, since the problem says that the equation has three distinct positive roots, and we have already used Vieta's formulas to get that product is 32, so maybe that's all we need. Therefore, a = -256.

But let me check if there's another way. Suppose that instead of Vieta's formula, we can relate the logarithms. Let me denote the roots as 2^m, 2^n, 2^p, so that log_2(r1) = m, log_2(r2) = n, log_2(r3) = p. Then, m + n + p = 5. Also, since r1 = 2^m, r2 = 2^n, r3 = 2^p, their product is 2^{m + n + p} = 2^5 = 32. Therefore, that's consistent. Therefore, the product of the roots is 32. Therefore, Vieta's formula gives us product as -a/8 = 32, so a = -256. Therefore, that seems correct.

But let me check if there's a possibility that the product is negative? But since all roots are positive, their product is positive. Therefore, a must be negative. Therefore, the answer is -256.

Wait, but let me check with actual numbers. Let me assume that a = -256. Let me suppose that the roots are 2^m, 2^n, 2^p, so that their product is 32. Let me pick some numbers. For example, if m + n + p = 5, and each term is positive. Let me assume that m, n, p are integers? Because if they are not integers, then 2^m, 2^n, 2^p would be real numbers, but maybe the problem doesn't require the roots to be integers. However, the problem says ``three distinct positive roots''. It doesn't specify they have to be integers. Therefore, even if they are not integers, as long as they are positive and distinct, that's okay.

But if I take m, n, p as real numbers such that m + n + p = 5, and m, n, p are real numbers. However, Vieta's formula gives me that the product is 32, which is positive. Therefore, even if m, n, p are real numbers, their product is positive. Therefore, regardless of the values, as long as they are positive, their product is positive. Therefore, the answer is indeed a = -256.

But let me check once more. Let me plug a = -256 into the original equation and see if it has three distinct positive roots. Let me take the equation 8x^3 + 4*(-256)x^2 + 2b x + (-256) = 0. Which simplifies to 8x^3 - 1024x^2 + 2b x - 256 = 0. Let me factor this equation. Let me see if I can factor it. Let me divide by 8: x^3 - 128x^2 + (b/4)x - 32 = 0. Hmm. Let me try to factor this cubic equation. Suppose it factors as (x - r1)(x - r2)(x - r3) where r1, r2, r3 are positive. Then, expanding:

(x - r1)(x - r2)(x - r3) = x^3 - (r1 + r2 + r3)x^2 + (r1r2 + r1r3 + r2r3)x - r1r2r3.

Comparing with the equation x^3 - 128x^2 + (b/4)x - 32 = 0, we get:

r1 + r2 + r3 = 128,

r1r2 + r1r3 + r2r3 = b/4,

and r1r2r3 = 32.

But since we have already used Vieta's formulas to get these values, and the problem gives us that the product is 32, so unless there's another condition, we can't find b. However, since the problem says that the equation has three distinct positive roots, and we have already used Vieta's formula to get that product, perhaps there's no more information? But then why does the problem specify that the sum of the logs is 5? Because if a = -256, then the product is 32, which is the sum of logs. Therefore, regardless of the value of b, as long as the roots multiply to 32, which they do, then the answer is correct.

Therefore, I think my answer is correct. The value of a is -256.

**Final Answer**
\boxed{-256}
</think>
\end{LLMBlock}

\FieldHeader{Response}
\begin{LLMBlock}
To solve the problem, we start with the cubic equation:

$$
8x^3 + 4ax^2 + 2bx + a = 0
$$

We are told that it has **three distinct positive roots**, and the **sum of the base-2 logarithms of the roots is 5**. Let the roots be $ r_1, r_2, r_3 $, then we can express the logarithmic sum as:

$$
\log_2(r_1) + \log_2(r_2) + \log_2(r_3) = \log_2(r_1 r_2 r_3) = 5
$$

This implies:

$$
r_1 r_2 r_3 = 2^5 = 32
$$

---

### Step 1: Use Vieta's Formulas

Let's rewrite the cubic equation in standard form to apply Vieta's formulas:

$$
x^3 + \left(\frac{a}{2}\right)x^2 + \left(\frac{b}{4}\right)x + \left(\frac{a}{8}\right) = 0
$$

From Vieta's formulas:

- Sum of roots: $ r_1 + r_2 + r_3 = -\frac{a}{2} $
- Sum of products two at a time: $ r_1 r_2 + r_1 r_3 + r_2 r_3 = \frac{b}{4} $
- Product of roots: $ r_1 r_2 r_3 = -\frac{a}{8} $

---

### Step 2: Use the Logarithmic Condition

We already established that:

$$
r_1 r_2 r_3 = 32
$$

But from Vieta's, this is also equal to:

$$
-\frac{a}{8}
$$

Therefore:

$$
-\frac{a}{8} = 32 \Rightarrow a = -256
$$

---

### Step 3: Verify the Roots

If $ a = -256 $, then the product of the roots is:

$$
-\frac{a}{8} = -\frac{-256}{8} = 32
$$

Which matches the condition that the product of the roots is $ 32 $. Since all roots are **positive**, this is consistent.

---

### Final Answer

$$
\boxed{-256}
$$
\end{LLMBlock}
\end{InteractionBox}

\begin{InteractionBox}{Native Non-Thinking Mode}

\FieldHeader{Problem}
\begin{LLMBlock}
<|im_start|>user
For some real numbers $a$ and $b$, the equation \[
8x^3 + 4ax^2 + 2bx + a = 0
\]has three distinct positive roots. If the sum of the base-2 logarithms of the roots is 5, what is the value of $a$?
\end{LLMBlock}

\FieldHeader{Prompt}
\begin{LLMBlock}
Please reason step by step, and put your final answer within \boxed{}.<|im_end|>
<|im_start|>assistant
<think>

</think>
\end{LLMBlock}

\FieldHeader{Response}
\begin{LLMBlock}
We are given the cubic equation:

$$
8x^3 + 4ax^2 + 2bx + a = 0
$$

and the condition:

$$
\text{Sum of the base-2 logarithms of the roots} = 5
$$

Let the roots be $ r_1, r_2, r_3 $, which are **distinct positive real numbers**.

---

### Step 1: Use the fact that the sum of the logarithms equals 5

Recall that:

$$
\log_2 r_1 + \log_2 r_2 + \log_2 r_3 = \log_2 (r_1 r_2 r_3)
$$

So:

$$
\log_2 (r_1 r_2 r_3) = 5 \Rightarrow r_1 r_2 r_3 = 2^5 = 32
$$

---

### Step 2: Use Vieta's formulas

We are given:

$$
8x^3 + 4ax^2 + 2bx + a = 0
$$

Let's write this in standard form:

$$
8x^3 + 4ax^2 + 2bx + a = 0
$$

Let's factor this as a cubic:

$$
(2x + 1)(4x^2 + px + q) = 0
$$

Expanding:

$$
2x(4x^2 + px + q) + 1(4x^2 + px + q) = 0
$$

$$
8x^3 + 2px^2 + 2qx + 4x^2 + px + q = 0
$$

Now group like terms:

- $ 8x^3 $
- $ (2p + 4)x^2 $
- $ (2q + p)x $
- Constant term $ q $

So:

- $ a = 2p + 4 $
- $ b = 2q + p $
- $ c = q $

Also, from Vieta's formulas:

- Sum of roots: $ r_1 + r_2 + r_3 = -\frac{4a}{8} = -\frac{a}{2} $
- Sum of products: $ r_1 r_2 + r_1 r_3 + r_2 r_3 = \frac{2b}{8} = \frac{b}{4} $
- Product of roots: $ r_1 r_2 r_3 = \frac{a}{8} $

But earlier we found:

- $ r_1 r_2 r_3 = 32 $

So:

$$
\frac{a}{8} = 32 \Rightarrow a = 256
$$

---

### Final Answer:

$$
\boxed{256}
$$
\end{LLMBlock}
\end{InteractionBox}

\begin{InteractionBox}{System~2 Distillation}

\FieldHeader{Problem}
\begin{LLMBlock}
<|im_start|>user
For some real numbers $a$ and $b$, the equation \[
8x^3 + 4ax^2 + 2bx + a = 0
\]has three distinct positive roots. If the sum of the base-2 logarithms of the roots is 5, what is the value of $a$?
\end{LLMBlock}

\FieldHeader{Prompt}
\begin{LLMBlock}
Please reason step by step, and put your final answer within \boxed{}.<|im_end|>
<|im_start|>assistant
\end{LLMBlock}

\FieldHeader{Response}
\begin{LLMBlock}
To solve the problem, we are given a cubic polynomial:

$$
8x^3 + 4ax^2 + 2bx + a = 0
$$

and the condition that it has **three distinct positive roots**. We are also told that the **sum of the base-2 logarithms of the roots is 5**. Our goal is to find the value of $a$.

---

### Step 1: Use Vieta's Theorem

Let the roots of the polynomial be $ r_1, r_2, r_3 $. Then:

- $ r_1 + r_2 + r_3 = -\frac{4a}{8} = \frac{a}{2} $
- $ r_1r_2 + r_1r_3 + r_2r_3 = \frac{2b}{8} = \frac{b}{4} $
- $ r_1r_2r_3 = \frac{a}{8} $

---

### Step 2: Interpret the Sum of Logarithms

The sum of the base-2 logarithms of the roots is given as:

$$
\log_2(r_1) + \log_2(r_2) + \log_2(r_3) = 5
$$

Using the identity:

$$
\log_2(r_1) + \log_2(r_2) + \log_2(r_3) = \log_2(r_1r_2r_3)
$$

We already know from Vieta's:

$$
r_1r_2r_3 = \frac{a}{8}
$$

So the sum of the logarithms is:

$$
\log_2\left(\frac{a}{8}\right) = 5
$$

---

### Step 3: Solve for $a$

$$
\log_2\left(\frac{a}{8}\right) = 5 \Rightarrow \frac{a}{8} = 2^5 = 32
$$

$$
a = 32 \times 8 = 256
$$

---

### Final Answer

$$
\boxed{256}
$$
\end{LLMBlock}
\end{InteractionBox}

\begin{InteractionBox}{\ours{}}

\FieldHeader{Problem}
\begin{LLMBlock}
<|im_start|>user
For some real numbers $a$ and $b$, the equation \[
8x^3 + 4ax^2 + 2bx + a = 0
\]has three distinct positive roots. If the sum of the base-2 logarithms of the roots is 5, what is the value of $a$?
\end{LLMBlock}

\FieldHeader{Prompt}
\begin{LLMBlock}
Please reason step by step, and put your final answer within \boxed{}.<|im_end|>
<|im_start|>assistant
\end{LLMBlock}

\FieldHeader{Response}
\begin{LLMBlock}
To solve the problem, we begin by analyzing the given cubic equation:

$$
8x^3 + 4ax^2 + 2bx + a = 0
$$

This equation is a cubic polynomial with real coefficients. The problem states that it has three **distinct positive roots**, and we are also given that the **sum of the base-2 logarithms of the roots is 5**. We are to find the value of $ a $.

---

### Step 1: Use Vieta's Formulas

Since the polynomial has three distinct positive roots, let the roots be $ r_1, r_2, r_3 $. By Vieta's formulas, we can relate the coefficients of the polynomial to the roots:

- The sum of the roots: $ r_1 + r_2 + r_3 = -\frac{4a}{8} = -\frac{a}{2} $
- The sum of the product of roots taken two at a time: $ r_1r_2 + r_1r_3 + r_2r_3 = \frac{2b}{8} = \frac{b}{4} $
- The product of the roots: $ r_1r_2r_3 = -\frac{a}{8} $

We are also told that:

$$
\log_2(r_1) + \log_2(r_2) + \log_2(r_3) = 5
$$

Using the logarithmic identity:

$$
\log_2(r_1) + \log_2(r_2) + \log_2(r_3) = \log_2(r_1 r_2 r_3)
$$

So:

$$
\log_2(r_1 r_2 r_3) = 5 \Rightarrow r_1 r_2 r_3 = 2^5 = 32
$$

---

### Step 2: Use the Product of the Roots

From Vieta's formula, we know:

$$
r_1 r_2 r_3 = -\frac{a}{8} = 32
$$

So:

$$
-\frac{a}{8} = 32 \Rightarrow a = -256
$$

---

### Final Answer

$$
\boxed{-256}
$$
\end{LLMBlock}
\end{InteractionBox}

\end{document}